%% file: main.tex
\documentclass{article}

\usepackage{arxiv}

\usepackage{times}
\usepackage{soul}
\usepackage{pifont}
\usepackage[utf8]{inputenc} 
\usepackage[T1]{fontenc}    
\usepackage{hyperref}       
\usepackage{url}            
\usepackage{graphicx}
\usepackage[small]{caption}
\usepackage{amsmath}
\usepackage{amsthm}
\usepackage[switch]{lineno}
\usepackage{longtable}
\usepackage{supertabular}
\usepackage{multirow}
\usepackage{tikz}                  
\usepackage{mathpazo} 
\usepackage{mdframed}

\usepackage[utf8]{inputenc} 
\usepackage[T1]{fontenc}    
\usepackage{url}            
\usepackage{booktabs}       
\usepackage{amsfonts}       
\usepackage{microtype}      
\usepackage{lipsum}         

\usepackage{graphicx}  
\usepackage{float} 
\usepackage{subfigure} 
\usepackage{algorithm}
\usepackage{algorithmic}
\usepackage{tabulary}
\usepackage{xspace}
\usepackage{natbib}

\usepackage{amssymb}
\usepackage{appendix}

\input{packages/defs}

\input{packages/header}

\input{packages/math_commands}
\newcommand{\piE}{\pi^{\operatorname{E}}}
\newcommand{\pirl}{\pi^{\operatorname{RL}}}

\newcommand{\name}{EndoRM\xspace}

\allowdisplaybreaks[4]

\date{}

\title{Generalist Reward Models: Found Inside Large Language Models 
}

\author{
  Yi-Chen Li$^{*}$, Tian Xu$^{*}$, Yang Yu$^{*,\dag}$, Xuqin Zhang, Xiong-Hui Chen, \\\textbf{Zhongxiang Ling, Ningjing Chao, Lei Yuan, Zhi-Hua Zhou}\\[3pt]
  National Key Laboratory for Novel Software Technology, Nanjing University, China\\
  School of Artificial Intelligence, Nanjing University, China \\
  $^*$ Equal contribution\\
  $^\dag$ Correspondence: \texttt{yuy@nju.edu.cn} \\
}

\begin{document}
\maketitle

\begin{abstract}
The alignment of Large Language Models (LLMs) is critically dependent on reward models trained on costly human preference data. While recent work explores bypassing this cost with AI feedback, these methods often lack a rigorous theoretical foundation. In this paper, we discover that a powerful generalist reward model is already latently present within any LLM trained via standard next-token prediction. We prove that this \emph{endogenous reward} is not a heuristic, but is theoretically equivalent to a reward function learned through offline inverse reinforcement learning. This connection allows us to directly elicit a high-quality reward signal from a base (pre-trained or supervised fine-tuned) model without any further training. Critically, we also prove that subsequent reinforcement learning using this endogenous reward leads to a policy with a provably superior error bound compared to the base model. To our best knowledge, this is the first theoretical proof of the effectiveness of reinforcement learning for LLMs. Our experiments validate this theory, demonstrating that our method not only outperforms existing LLM-as-a-judge approaches but can also surpass explicitly trained reward models. 

These findings suggest that the reward modeling stage can be replaced by a principled method of eliciting the knowledge already captured during pre-training, heralding a more efficient, powerful, and scalable paradigm for LLMs alignment as well as multi-modal models.
\end{abstract}

\section{Introduction}
Aligning Large Language Models (LLMs) with complex human values, such as helpfulness and honesty, remains a central challenge in AI development. The predominant paradigm for this task is Reinforcement Learning from Human Feedback (RLHF)~\citep{christiano2017deep, bai2022training,openai2022chatgpt}. This pipeline critically relies on a reward model (RM) trained on human preferences to score model outputs, and the quality of the final, aligned LLM is fundamentally determined by the quality of this RM. Consequently, creating a state-of-the-art RM has previously required building massive, high-quality human preference datasets, which is notoriously slow, expensive, and difficult to scale.

This dependency on human labeling has spurred research into alternative alignment methods. A prominent direction is Reinforcement Learning from AI Feedback (RLAIF) \citep{bai.constitutionalai,pang2023yang,RLAIF} or the ``LLM-as-a-judge'' framework \citep{zheng2023judging,judge}, which uses a powerful proprietary LLM to generate reward signals or preference labels, thus bypassing the need for human annotators. While cost-effective, these methods are often \emph{heuristic}; they lack a rigorous theoretical foundation and can be susceptible to inheriting the stylistic quirks and biases of the judge model. This raises a critical question: must a high-quality reward signal be sourced externally at all?

This paper discover that a powerful generalist reward model is not something that needs to be built, but rather something that can be \emph{uncovered}, as it is already latently present within any language model trained via standard next-token prediction. We term this the \textbf{endogenous reward}.

Our core contribution is to provide a rigorous theoretical basis for this claim. We demonstrate that the reward function from a specific form of offline inverse reinforcement learning (IRL) can be recovered from the standard next-token prediction objective, used for both pre-training and supervised fine-tuning (SFT). This insight allows us to move beyond heuristics and establish a principled method for eliciting the reward function implicitly learned by the language model during its training. Specifically, we show that a language model's logits can be directly interpreted as the soft Q-function, from which a reward function can be recovered via the inverse soft Bellman operator.

Crucially, this theoretical connection provides more than just a method for reward extraction. We prove that fine-tuning a model with its own endogenous reward leads to a policy with a provably superior error bound compared to the base model. The RL process effectively corrects for the \emph{compounding errors} that plague standard imitation learning (i.e., next-token prediction), reducing the performance gap from a quadratic dependence on the task horizon, $\gO (H^2)$, to a superior linear one, $\gO (H)$. To our best knowledge, this is the first theoretical proof of the effectiveness of reinforcement learning for LLMs.

Our extensive experiments validate this theory, showing that eliciting this endogenous reward is not only superior to existing LLM-as-a-judge methods but can also surpass the performance of reward models trained explicitly on expensive human-labeled data. These findings imply that the reward modeling stage can potentially be replaced by a principled process of eliciting the knowledge already captured by the base model, heralding a more efficient, powerful, and theoretically grounded paradigm for LLM alignment.

\section{Preliminaries}

\subsection{LLM and its MDP Formulation}

\paragraph{Basic Introduction on LLM.} A Large Language Model (LLM) is a generative model that operates by predicting the next token in a sequence by probabilistic modeling. Formally, an LLM $\pi$ selects tokens from a finite vocabulary $\gV = \{1,2, \ldots, |\gV| \}$ and generates a sequence in an autoregressive manner. Specifically, in step $h$, given a context sequence $(a_1, \ldots, a_{h-1})$, an LLM generates the next token according to the conditional distribution, i.e., $a_h \sim \pi (\cdot|a_1, \ldots, a_{h-1})$. This autoregressive generation procedure proceeds until a designated End-Of-Sentence (EOS) token is generated or a predetermined maximum length $H$ is reached. For analytical clarity, we assume uniform response lengths of exactly $H$, with appropriate padding applied after the EOS token.

\paragraph{MDP Formulation of LLM.} In this paper, we take an reinforcement learning perspective of LLMs and formalize the language generation task as an MDP \citep{llm_formalization}, defined by a tuple $\mathcal{M} = \langle \mathcal{S}, \gV, r, P, \rho, H\rangle$. Here the state space $\gS$ is the set of all finite-length strings formed by the concatenation of elements in $\gV$ and the action space is the vocabulary set $\gV$. At the beginning of each trajectory, a prompt $s_1 = (x_1, x_2, \cdots, x_m)$ of length $m$ is sampled from the initial state distribution $\rho$, with $m \in \mathbb{N}$ and $\forall i \in [m], x_i \in \mathcal{V}$. At each step $h \in [H]$, the LLM selects an action (or equivalently, a token) $a_h \in \mathcal{V}$ according to $\pi(\cdot|s_h)$. The environment then transits to the next state $s_{h+1} = (x, a_1, \cdots, a_h)$, rewarding the LLM with $r(s_h, a_h) \in [0, 1]$. That is, the transition model $P: \mathcal{S}\times\mathcal{V}\to \Delta (\mathcal{S})$ is deterministic. $ P(s_{h+1}|s_h,a_h)=1$ if and only if $s_{h+1} = s_h \oplus a_h$, where $\oplus$ means concatenation. The trajectory ends after a total of $H$ steps. In the context of RL, we also call $\pi$ as a policy. Throughout this paper, the terms ``policy'' and ``LLM'' will be used interchangeably. We measure the quality of a policy $\pi$ by its policy value $V^{\pi}$ defined as $V^{\pi} = \expect_{\tau \sim \pi} \ls \sum_{h=1}^H r (s_h, a_h) \rs$.

\subsection{Next-token Prediction for Training LLMs}
Next-token prediction \citep{vaswani2017attention} serves as the fundamental training objective for LLMs, where the goal is to train an LLM to maximize the likelihood of predicting the next token given a sequence of preceding tokens. Formally, given the dataset $\gD =\{ (s^i_1, a^i_{1:H}) \}_{i=1}^n$, next-token prediction solves the following optimization problem. 
\begin{align}
\label{eq:next_token_prediction}
    \max_{\pi} \sum_{i=1}^n \sum_{h=1}^H \log \lp \pi \lp a^i_h |s^i_{h} \rp \rp.
\end{align}
This objective is employed across multiple LLMs training stages: during pre-training \citep{radford2019language, brown2020language} with massive general-purpose web datasets, and during supervised fine-tuning \citep{trainlmfollowinstruct} with smaller collections of high-quality human responses. Notably, next-token prediction can be viewed as an instance of the famous behavior cloning approach \citep{Pomerleau91bc} from an imitation learning (IL) perspective \citep{osa2018survey}, where the model mimics the expert's action in states from demonstrations.

\subsection{Reinforcement Learning from Human Feedback}
\label{subsec:rlhf}

While next-token prediction is effective for teaching a model to imitate high-quality demonstrations, aligning its behavior with human values often requires a more direct form of feedback. Reinforcement Learning from Human Feedback (RLHF)~\citep{trainlmfollowinstruct} is the standard paradigm for this task, which refines an SFT model using a reward signal derived from human preferences. The process typically involves two subsequent stages: reward modeling and RL fine-tuning.

\paragraph{Reward Modeling.}
The first step is to train a separate reward model (RM), denoted $R_\phi$ with parameters $\phi$, to predict which responses humans would prefer. This requires a preference dataset, $\mathcal{D}_{\text{pref}} = \{(x, y_w, y_l)_i\}$, where for a given prompt $x$, $y_w$ is a response preferred by a human annotator (the ``winner") and $y_l$ is the dispreferred response (the ``loser"). The RM is trained to assign a higher scalar score to the winning response. Following the Bradley-Terry (BT) model~\citep{bradley1952rank}, the probability that $y_w$ is preferred over $y_l$ is modeled as:
\begin{equation}
P(y_w \succ y_l | x) = \sigma(R_\phi(x, y_w) - R_\phi(x, y_l))
\end{equation}
where $\sigma$ is the sigmoid function. The RM's parameters $\phi$ are optimized by minimizing the negative log-likelihood of the observed preferences in the dataset:
\begin{equation}
\label{eq:preference_rm}
\mathcal{L}_{\text{RM}}(\phi) = -\mathbb{E}_{(x, y_w, y_l) \sim \mathcal{D}_{\text{pref}}} \left[ \log \sigma(R_\phi(x, y_w) - R_\phi(x, y_l)) \right]
\end{equation}

\paragraph{Reinforcement Learning Fine-Tuning.}
Once the RM is trained, the SFT model, $\pi_{\text{SFT}}$, is further optimized using reinforcement learning. The SFT model is treated as the initial policy, $\pi_\theta$, which is then updated to maximize the expected reward assigned by the RM. To prevent the policy from deviating too far from the distribution learned during SFT and to mitigate ``reward hacking'' (i.e., finding adversarial ways to maximize the reward score), a KL-divergence penalty is added to the objective. The final optimization problem is:
\begin{equation}
\label{eq:rlhf_objective}
\max_{\theta} \mathbb{E}_{x \sim \mathcal{D}, y \sim \pi_{\theta}(\cdot|x)} \left[ R_\phi(x, y) - \beta \cdot \KL (\pi_{\theta}(\cdot|x) \parallel \pi_{\text{SFT}}(\cdot|x)) \right]
\end{equation}
where $\mathcal{D}$ is a distribution of prompts, $y$ is the full response generated by the policy $\pi_\theta$, and $\beta$ is a hyperparameter controlling the strength of the KL penalty.

\subsection{Inverse Reinforcement Learning}
\label{subsec:irl}

Inverse Reinforcement Learning (IRL)~\citep{ng2000algorithms} is a fundamental class of imitation learning methods that addresses the inverse problem of RL. Instead of learning a policy from a known reward function, IRL seeks to recover a reward model from expert demonstrations, under the assumption that these demonstrations are optimal or near-optimal.

A prominent and theoretically grounded approach is Maximum Entropy (MaxEnt) IRL~\citep{ziebart2008maximum}. It resolves the ambiguity that many reward functions can explain the same behavior by seeking a reward that rationalizes the expert demonstrations while being maximally non-committal about the behavior uncovered in the data. This principle leads to a minimax optimization problem:
\begin{align} \label{eq:maxent_irl_objective}     
\max_{r} \min_{\pi} \left( \mathbb{E}_{\tau \sim \piE} \ls \sum_{h=1}^H r(s_h,a_h) \rs - \mathbb{E}_{\tau \sim \pi} \left[ \sum_{h=1}^H r(s_h, a_h) + \alpha H(\pi(\cdot|s_h)) \right] \right)
\end{align}
Here, the expectation $\mathbb{E}_{\tau \sim \piE}$ is taken over the trajectory distribution induced by the expert policy $\piE$ (approximated by the dataset $\mathcal{D}$), while the expectation $\mathbb{E}_{\tau \sim \pi}$ is over the trajectory distribution generated by a learned policy $\pi$. Besides, $H (\pi (\cdot|s_h)) = \expect_{a_h \sim \pi (\cdot|s_h)} \ls - \log \pi (a_h|s_h) \rs$ represents the entropy of $\pi (\cdot|s_h)$ and $\alpha > 0$ is the regularization coefficient. The objective seeks a reward function $r$ that maximizes the margin between the expert's expected return and the entropy-regularized return of the optimal policy $\pi$ under the same reward $r$.

\paragraph{Connection to Reward Modeling in RLHF.}
It is insightful to view the MaxEnt IRL objective as a more general, distribution-level formulation of the principle underlying the BT model used in RLHF. The objective under the BT model in Eq.(\ref{eq:preference_rm}) learns a reward function by maximizing the score difference between a specific ``winner'' trajectory $y_w$ and a ``loser'' trajectory $y_l$. This is fundamentally a \emph{pairwise} or \emph{sample-level} comparison.

The MaxEnt IRL objective in Eq. (\ref{eq:maxent_irl_objective}) can be seen as the \emph{distribution-level} analogue. Instead of comparing two sampled trajectories, it compares the entire expert distribution ($\piE$) against the optimal competing policy $\pi$ that arises under the reward function $r$. The `$\min_\pi$` term actively seeks the strongest possible competitor policy, and the `$\max_r$` term then adjusts the reward to create the largest possible performance gap.

In this light, the RLHF reward modeling process can be interpreted as a practical, computationally tractable instantiation of the IRL principle. Rather than solving the complex `$\min_\pi$` optimization problem, RLHF simplifies it and directly maximizes the reward margin between the ``winner'' against the ``loser'' in the preference dataset. Therefore, the BT model is a specific case within the broader theoretical framework that IRL provides. Our work returns to this more general IRL formulation to derive a reward function directly, without the need for explicit pairwise comparisons.

\section{Finding the Endogenous Reward Within the Language Model}
\label{sec:main_theory}

In this section, we establish our central theoretical claim: a principled reward function can be directly recovered from the logits of any language model trained with standard next-token prediction.

\subsection{Reward Learning from Inverse Reinforcement Learning}
\label{subsec:ideal_reward_irl}

As discussed in our preliminaries, the standard practice of reward modeling in RLHF can be seen as a practical, simplified instance of a more fundamental paradigm: Inverse Reinforcement Learning (IRL). This connection suggests that IRL is the natural and most principled framework for learning a reward function from data. This begs the question: rather than relying on the pairwise comparison heuristics common in RLHF, can we apply a more foundational IRL approach to recover the optimal reward function that best explains an expert dataset, $\mathcal{D}$? The primary challenge that has historically hindered this direct approach is that most IRL methods are formulated for online settings, requiring computationally expensive interactions.

A more direct and practical approach is to formulate the problem using an offline IRL method based on inverse soft Q-learning~\citep{garg2021iq-learn}. This method seeks to find a Q-function that best explains the expert data in a static dataset $\mathcal{D} =\{ (s^i_1, a^i_{1:H}) \}_{i=1}^n$. The objective is to solve the following optimization problem:
\begin{align}
\label{eq:offline_irl}
    \max_{Q} \frac{1}{n} \sum_{i=1}^n \sum_{h=1}^H \left[ Q (s^i_h, a^i_h) - \alpha \log \bigg( \sum_{a_{h} \in \gV} \exp \lp Q (s^i_{h}, a_{h}) \rp / \alpha \bigg) \right]
\end{align}
Once we find the optimal Q-function, $Q^\star$, that maximizes this objective, the corresponding ideal reward function, $r^\star$, can be recovered via the \emph{inverse soft Bellman operator}~\citep{garg2021iq-learn}:
\begin{align}
\label{eq:inverse_bellman_operator}
    r^\star(s_h, a_h) := Q^\star(s_h, a_h) - \alpha \log \lp \sum_{a_{h+1} \in \gV} \exp \lp Q^\star(s_{h+1}, a_{h+1}) \rp / \alpha \rp. 
\end{align}
This provides a principled, offline method to find a reward function. The critical question is: what is the solution, $Q^\star$, to this problem, and can we find it without a complex, new optimization procedure?

\subsection{Next-Token Prediction Recovers the IRL Solution}
\label{subsec:ntp_recovers_solution}

We now demonstrate that the solution to $Q^\star$ posed in Eq.~\eqref{eq:offline_irl} is not something that needs to be computed from scratch. Rather, it is already embodied by the logits of any language model trained with the standard next-token prediction objective.

To see this, let's first analyze what kind of solution the IRL objective seeks. By applying a simple log-of-exponent transformation, the objective in Eq.~\eqref{eq:offline_irl} can be expressed as a maximum likelihood problem:
\begin{align}
\label{eq:equivalence}
    \max_{Q} \frac{1}{n} \sum_{i=1}^n \sum_{h=1}^H \log \lp \frac{\exp ( Q(s^i_h, a^i_h) / \alpha)}{\sum_{a'_{h} \in \gV} \exp ( Q(s^i_h, a'_{h}) / \alpha)  } \rp.
\end{align}
The term inside the logarithm can be regarded as a policy, $\pi_Q(\cdot|s_h) = \softmax(Q(s_h, \cdot); \alpha)$. Therefore, the optimization problem seeks the Q-function whose corresponding policy, $\pi_Q$, maximizes the likelihood of the expert demonstrations in dataset $\mathcal{D}$.

This is precisely the goal of training a language model by next-token prediction. During its training, a language model, $\widehat{\pi}$, is optimized to maximize the log-likelihood of the data as shown in Eq.(\ref{eq:next_token_prediction}). The resulting policy, $\widehat{\pi}$, is by definition the maximum likelihood estimate of the data-generating distribution. This policy is parameterized as a softmax over the model's logits, which we denote as $\widehat{f}$:
\begin{equation}
    \widehat{\pi}(\cdot|s_h) = \softmax(\widehat{f}(s_h, \cdot); \alpha).
\end{equation}
Since the language model's policy $\widehat{\pi}$ is the solution to the maximum likelihood problem, it directly follows that its underlying logits, $\widehat{f}$, are a valid solution for the Q-function in the IRL optimization problem of Eq.~\eqref{eq:offline_irl}. We formalize this direct connection as follows:

\begin{prop}
\label{prop:from_ntp_to_irl}
    Let $\widehat{\pi}$ be a language model trained on a dataset $\mathcal{D}$ via next-token prediction, yielding a policy $\widehat{\pi}(\cdot|s_h) = \softmax(\widehat{f}(s_h, \cdot); \alpha)$, where $\widehat{f}$ are the model's logits. The logits function $\widehat{f}$ is a solution to the principled offline IRL objective in Eq.~\eqref{eq:offline_irl}.
\end{prop}

This proposition provides the theoretical cornerstone of our work. It reveals that the logits of a language model are not just arbitrary scores; they are a principled Q-function that implicitly represents the optimal reward function for the data the model was trained on. This discovery provides a unified view on generation and evaluation, where the model's policy $\widehat{\pi}$ is responsible for generation, while its logits $\widehat{f}$ (acting as a Q-function) are responsible for evaluation.

\subsection{The Endogenous Reward}
\label{subsec:uncovering_reward}

The direct connection established in \cref{prop:from_ntp_to_irl} gives us a powerful, training-free method to obtain a reward model. Given any language model trained via next-token prediction, whether during pre-training or a fine-tuning stage, we can simply take its logits, $\widehat{Q} = \widehat{f}$, and substitute them into the inverse soft Bellman operator from Eq.~\eqref{eq:inverse_bellman_operator}. 
\begin{align}
\label{eq:extracted_reward}
    \widehat{r} (s_h, a_h) = \widehat{Q} (s_h, a_h) - \alpha \log \lp \sum_{a_{h+1} \in \gV} \exp \lp \widehat{Q} (s_{h+1}, a_{h+1}) \rp / \alpha \rp. 
\end{align}
Let us define the value function $V_{\widehat{Q}}(s_h) := \alpha \log ( \sum_{a_h \in \gV} \exp ( \widehat{Q} (s_h, a_h) / \alpha))$, which represents the expected future returns from state $s$. This results in the \textbf{endogenous reward}, written as:
\begin{align}
\label{eq:reward_log_policy}
    \widehat{r} (s_h, a_h) = \alpha \log \lp \widehat{\pi} (a_h|s_h) \rp + V_{\widehat{Q}} (s_h) - V_{\widehat{Q}} (s_{h+1}).
\end{align}
We elaborate on this endogenous reward in the following three aspects.
    
\paragraph{Reward Shaping.} 
From Eq.\eqref{eq:reward_log_policy}, careful readers may notice that the endogenous reward $\widehat{r}$ can be viewed as a shaped reward \citep{ng1999policy} of $\widetilde{r} (s_h, a_h) := \log (\widehat{\pi} (a_h|s_h) )$ represented by the language model's log-probability, with the potential function $V_{\widehat{Q}}$. According to the theory on reward shaping \citep{ng1999policy}, $\widehat{r}$ and $\widetilde{r}$ induce the same optimal policy and value function.

\paragraph{Outcome Reward.} For a complete response $\tau = (s_1, a_1, \ldots, s_H, a_H)$, the resultant outcome reward can be calculated as 
\begin{align}
\label{eq:outcome_reward}
    \widehat{r} (\tau) = \sum_{h=1}^H \widehat{r} (s_h, a_h) = \alpha \sum_{h=1}^H \log \lp \widehat{\pi} (a_h|s_h) \rp + V_{Q}(s_1) = \alpha \log (\widehat{\pi} (\tau|s_1)) + V_{\widehat{Q}} (s_1).  
\end{align}
The second equation follows the telescoping argument with the boundary condition $\widehat{Q} (s_{H+1}, \cdot) \equiv 0$. $\widehat{\pi} (\tau|s_1) := \prod_{h=1}^H \widehat{\pi} (a_h|s_h)$ represents the probability of the complete response $\tau$ under $\widehat{\pi}$. We can observe that the outcome reward equals the sum of log probabilities across the trajectory, up to the term $V_{Q} (s_1)$ only depending on the initial state (i.e., prompt). Intuitively, if a response occurs frequently in the training data (e.g., massive pre-training data), then the trained LLM will generate this response with a higher likelihood, which also means the reward value of this response is higher.

\paragraph{Exisiting Generative Reward Models as Special Cases.}  Eq.(\ref{eq:outcome_reward}) introduces a general formulation that exploits a LLM's probability for reward construction, with existing generative reward models \citep{mahan2024generative, zhang2024generative} emerging as particular instances of this framework. For instance, for a question-answer pair $(\textbf{x}, \textbf{y})$, the generative verifier in \citep{zhang2024generative} uses a prompt $\textbf{P}=\text{`Is the answer correct?'}$, and represents the reward as the probability of a single token ‘Yes’ under the context and the prompt, i.e., $\widehat{\pi} (\text{Yes}| \textbf{x}, \textbf{y}, \textbf{P})$. This corresponds to our method when setting $s_1 = (\textbf{x}, \textbf{y}, \textbf{P})$ and $a_1 = \text{`Yes'}$, resulting the reward $\widehat{r} (\tau) = \alpha \log ( \widehat{\pi} (\text{Yes}| \textbf{x}, \textbf{y}, \textbf{P}) ) $ consistent with theirs. Similar arguments can also be made for the generative reward model approach in \citep{mahan2024generative} designed for pair-wise comparisons. Based on this connection, the established \cref{prop:from_ntp_to_irl} can also provide a theoretical justification for these approaches.


\section{Theoretical Justifications for the Endogenous Reward}
In this part, we provide theoretical justifications for the endogenous reward by analyzing both the reward error and the induced policy error.

\subsection{Error Analysis of the Endogenous Reward.} In the sequel, we analyze the performance of the endogenous reward. Specifically, following maximum entropy inverse reinforcement learning \citep{ziebart2008maximum}, we consider that the expert policy $\piE$ is the entropy-regularized optimal policy regarding the unknown true reward $r^\star$. According to the theory on regularized MDPs \citep{geist2019theory}, it holds that
\begin{align}
\label{eq:true_reward_log_expert}
    r^\star (s_h, a_h) = \alpha \log \lp \piE (a_h|s_h) \rp + V_{Q^{\star}_{r^\star}} (s_h) - V_{Q^\star_{r^\star}} (s_{h+1}).
\end{align}
Here $Q^\star_{r^\star}$ is the entropy-regularized optimal Q-value function with the true reward $r^\star$. First, due to the reward ambiguity issue that multiple reward functions can derive the same expert policy \citep{ng1999policy}, it is impossible to approximate the true reward $r^\star$ even with the full knowledge of $\piE$. Therefore, we cannot draw any theoretical claim on the absolute error $| \widehat{r} (s_h, a_h) - r^\star (s_h, a_h)  |$.

Nevertheless, recall one primary goal of building reward models is to perform pair-wise comparisons \citep{trainlmfollowinstruct, bai2022training, malik2025rewardbench}. Therefore, instead of analyzing the absolute error $| \widehat{r} - r^\star |$, we turn to analyze the performance when leveraging reward models to perform comparisons for two responses. 
Formally, the task is to leverage the reward model to evaluate the preference over a pair of responses $(\tau, \tau^\prime)$ with $\tau = (s_1, a_{1:H}), \tau^\prime = (s_1, a^\prime_{1:H})$. Following the Bradley-Terry model \citep{bradley1952rank}, the preference distribution induced by reward $r$ can be formulated as $\sP_{r} (\tau \succ \tau^\prime |\tau, \tau^\prime) = \sigma (r (\tau) - r (\tau^\prime) )$, where $\sigma (x) =1/ (1+\exp (-x))$ is the sigmoid function. The theoretical goal is to analyze the divergence between the true preference distribution $\sP_{r^\star} (\cdot|\tau, \tau^\prime)$ and the inferred one $\sP_{\widehat{r}} (\cdot|\tau, \tau^\prime)$.



\begin{thm}
    \label{thm:bound_on_preference_distribution}
    In token-level MDPs with the unknown true reward $r^\star$, suppose that the expert policy $\piE$ is the entropy-regularized optimal policy. Consider that $\widehat{\pi}$ is the policy trained by next-token prediction and $\widehat{r}$ is the endogenous reward defined in Eq.(\ref{eq:extracted_reward}). For any pair of responses $(\tau, \tau^\prime)$ with $\tau = (s_1, a_{1:H}), \tau^\prime = (s_1, a^\prime_{1:H})$, we have that
    \begin{align*}
        D_{\TV} \lp \sP_{r^\star} (\cdot |\tau, \tau^\prime), \sP_{\widehat{r}} (\cdot|\tau, \tau^\prime)   \rp \leq \frac{\alpha H}{2} \varepsilon_{\pi}. 
    \end{align*}
    Here $D_{\TV} (p, q) = (1/2) \sum_{x \in \gX} | p(x) - q (x)|$ is the total variation distance between two distributions $p, q \in \Delta (\gX)$ and $\varepsilon_{\pi} := \max_{h \in [H]} \max_{(s_h, a_h)} | \log (\piE (a_h|s_h)) - \log (\widehat{\pi} (a_h|s_h))  | $.
\end{thm}
\cref{thm:bound_on_preference_distribution} indicates that if the LLM $\widehat{\pi}$ used to extract rewards is closed to the underlying expert $\piE$ in log-probabilities across the responses, then the preference distribution induced by $\widehat{r}$ is closed to that induced by $r^\star$. This suggests that the proposed endogenous reward extracted from LLMs can provably inherit their performance.


\subsection{Error Analysis of the LLM Finetuned by the Endogenous Reward} 
The ultimate goal of eliciting the reward function is to use it for training a new, improved policy via reinforcement learning. Here we analyze the performance of the newly learned policy and make a comparison with the base policy.

Specifically, we analyze two approaches. The first approach directly applies next-token prediction (i.e., behavior cloning) on demonstrations, yielding the base policy $\widehat{\pi}$. Another approach first builds the endogenous reward based on Eq.(\ref{eq:extracted_reward}) from $\widehat{\pi}$, and then applies reinforcement learning using $\widehat{r}$, yielding the new policy $\pirl = \argmax_{\pi} V^{\pi}_{\widehat{r}}$. Here, we ignore any optimization error in solving the optimal policy. The theoretical goal is to analyze and compare the sub-optimality of $\pirl$ and $\widehat{\pi}$ under the true reward $r^\star$.

\begin{thm}
\label{thm:policy_error}
    In token-level MDPs with the unknown true reward $r^\star$, suppose that the expert policy $\piE$ is the entropy-regularized optimal policy. Consider that $\widehat{\pi}$ is the policy trained by next-token prediction and $\widehat{r}$ is the extracted reward defined in Eq.(\ref{eq:extracted_reward}). Apply reinforcement learning using the extracted reward $\widehat{r}$ to learn a new policy $\pirl$, i.e., $\pirl =\argmax_{\pi} V^{\pi}_{\widehat{r}}$. Then we have that
    \begin{align*}
        V^{\piE}_{r^\star} - V^{\widehat{\pi}}_{r^\star}  \precsim H^2 \varepsilon_{\pi}
        \qquad\qquad \textrm{and}\qquad\qquad  V^{\piE}_{r^\star} - V^{\pirl}_{r^\star}  \precsim  H \varepsilon_{\pi}. 
    \end{align*}
    Here $V^{\pi}_{r}$ represents the value of $\pi$ under $r$, $\varepsilon_{\pi} := \max_{h \in [H]} \max_{(s_h, a_h)} | \log (\piE (a_h|s_h)) - \log (\widehat{\pi} (a_h|s_h))  | $.
\end{thm}
\cref{thm:policy_error} indicates that $\pirl$ achieves a linear dependence on the response length $H$ in the sub-optimality bound, while $\widehat{\pi}$ suffers a quadractic dependence. This quadratic dependence reflects the compounding errors issue in imitation learning \citep{ross2010efficient, xu2021error}, where the single-step error $\varepsilon_{\pi}$ accumulates along with the trajectory. The superior linear dependence achieved by $\pirl$ reveals a fundamental advantage of inverse reinforcement learning. Instead of directly imitating expert actions, recovering the underlying reward function and performing RL with this reward eliminates the compounding error problem. As a technical remark, although this paper focuses on the finite-horizon setup, \cref{thm:policy_error} can be extended to the infinite-horizon setting through the analysis developed in \citep{xu2021error}.       

\subsection{Ineffectiveness of Iterative Improvement}
A natural question is whether this self-improvement process can be repeated: can we take the improved policy $\pirl$, extract its own endogenous reward, and perform another RL step to gain further improvements? The answer is no. By construction, the policy $\pirl$ is the optimal policy for the reward $\widehat{r}$. Consequently, the endogenous reward extracted from $\pirl$ is precisely the one for which $\pirl$ is already optimal. Applying another RL step would therefore yield the same policy, causing the process to converge immediately.

\section{Experiments}\label{sec:exp}

In this section, we conduct experiments to empirically validate our theoretical claims\footnote{Due to time limitation, we report initial experimental results, and more results will be presented in future versions.}. Our evaluation is designed to answer three central research questions:
\begin{itemize}
    \item \textbf{Q1:} How does the training-free endogenous reward model (\name) perform on common reward model benchmarks compared to both heuristic baselines and explicitly trained, state-of-the-art reward models?
    \item \textbf{Q2:} Does the endogenous reward have a strong instruction-following ability, functioning as a generalist reward model that can be prompted?
    \item \textbf{Q3:} Can reinforcement learning with the endogenous reward yield a better policy, achieving self-improvement as predicted by our theory?
\end{itemize}

\subsection{Experimental Setup}

\paragraph{Datasets} We use \textbf{RM-Bench}~\citep{rm-bench} as a testbed to benchmark the general performance of different reward models (\textbf{Q1}). To evaluate the instruction-following ability of \name (\textbf{Q2}), we use the \textbf{Multifaceted-Bench}~\citep{mf-bench}, which contains diverse user-specific preferences, and the \textbf{Domain-Specific Preference (DSP)} dataset~\citep{dsp}, which has distinct preferences for different professional domains. Finally, to verify that RL with our \name can lead to self-improvement (\textbf{Q3}), we use \textbf{MATH-lighteval}~\citep{hendrycks2measuring} as the training dataset.

\paragraph{Evaluation Metrics} For RM-Bench, Multifacted-Bench and DSP, we calculate the classification accuracy of the reward model for the two responses under a given prompt. For our method, we compute the total reward sum over all tokens for each response, as shown in \cref{exp:outcome}. The response with the higher reward sum is classified as "chosen". For RL, we select five widely used math reasoning benchmarks including AIME 2024, AMC~\citep{li2024numinamath}, Minerva~\citep{DBLP:journals/corr/abs-2206-14858}, OlympiadBench~\citep{he2024olympiadbench}, MATH-500~\citep{hendrycks2measuring}. The temperatures for training and test are set to 1.0 and 0.6, respectively. For AIME 2024 and AMC, we report avg@32, and for the other three benchmarks, we report pass@1.

\subsection{Reward Accuracy on Diverse Preference Pairs (Q1)}

To answer \textbf{Q1}, we evaluate reward models by using them to predict the chosen response in RM-Bench. Higher accuracy implies better quality of the reward. Since our method does not require training, we compare it against other training-free approaches: \textbf{Generative Verifier}~\citep{zhang2024generative}, \textbf{GenRM-Pairwise}, and \textbf{GenRM-Pointwise}~\citep{mahan2024generative}. The prompt templates are in Appendix~\ref{exp:prompt}. All baselines and our \name use \textbf{Qwen2.5-7B-Instruct}~\citep{qwen2.5} as the base model for a fair comparison. We also include results from four top-performing, explicitly trained reward models for context.

The results in \cref{table:rm-bench} show that \name not only substantially outperforms all training-free baselines using the same base model but also achieves a higher average score than the state-of-the-art explicitly trained reward models, supporting that the endogenous reward can be more effective than reward models that require expensive preference data curation and training. 

\begin{figure}[h!]
\begin{center}
    \includegraphics[width=1\textwidth]{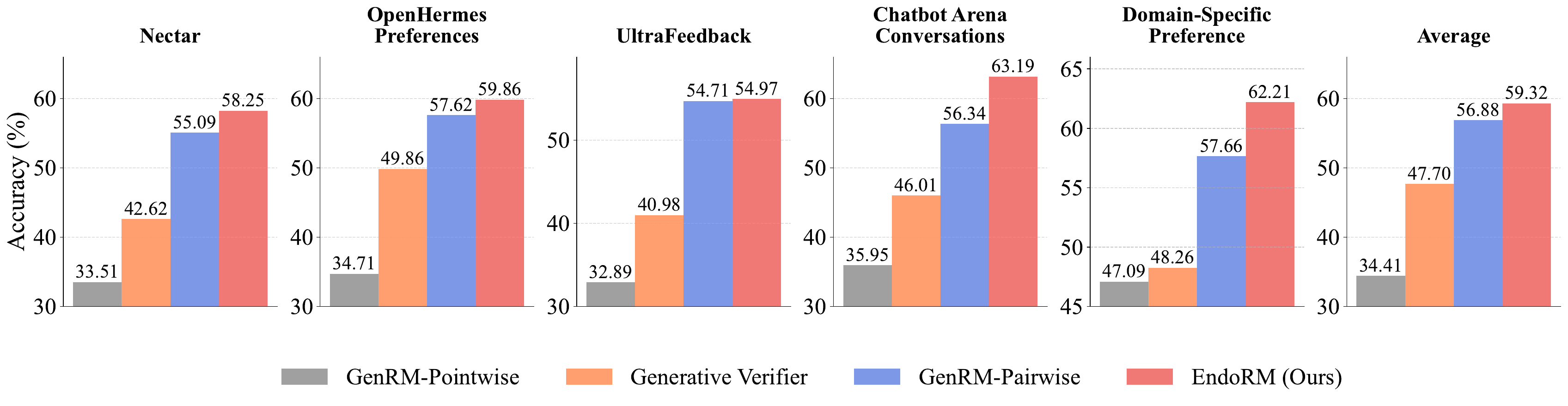}
\end{center}
\caption{Response classification accuracy on the Multifacted-Bench.}
\label{fig:mfb-result}
\end{figure}

We further present the results on the Multifaceted-Bench in \cref{fig:mfb-result}, from which we can see that our \name consistently outperforms all the baselines across five domains. Considering that there can be thousands of preferences in Multifaceted-Bench, this demonstrates that \name achieves scalable robustness even as task complexity and preference diversity increase. This supports our core hypothesis that a powerful reward signal is already latent within the base model.




\begin{table}[h!]
    \caption{Response classification accuracy by reward models on the RM-Bench. We bold highest score in each column.}
    \vspace{0.5em} 
    \centering
    \setlength{\tabcolsep}{3.3pt} %
    \scalebox{0.92}{
    \begin{tabular}{l|cccc|ccc|c}
    \toprule
        Model Name & \thead{Chat} & \thead{Math} & \thead{Code} & \thead{Safety} & \thead{Easy} & \thead{Normal} & \thead{Hard} & \thead{Avg} \vspace{-3pt} \\ 
        \midrule
        \href{https://huggingface.co/Skywork/Skywork-Reward-Llama-3.1-8B}{ Skywork/Skywork-Reward-Llama-3.1-8B} & $ 69.5 $ & $ \textbf{60.6} $ & $ 54.5 $ & $ \textbf{95.7} $ & $ \textbf{89.0} $ & $ \textbf{74.7} $ & $ 46.6 $ & $ 70.1 $ \\
        \href{https://huggingface.co/nvidia/Nemotron-4-340B-Reward}{\ NVIDIA/Nemotron-340B-Reward} & $ 71.2 $ & $ 59.8 $ & $ \textbf{59.4} $ & $ 87.5 $ & $ 81.0 $ & $ 71.4 $ & $ 56.1 $ & $ 69.5 $ \\
        \href{https://huggingface.co/Ray2333/GRM-llama3-8B-distill}{ Ray2333/GRM-llama3-8B-distill} & $ 62.4 $ & $ 62.1 $ & $ 56.9 $ & $ 88.1 $ & $ 82.2 $ & $ 71.5 $ & $ 48.4 $ & $ 67.4 $ \\
        \href{https://huggingface.co/allenai/tulu-2-dpo-13b}{ allenai/tulu-2-dpo-13b} & $ 66.4 $ & $ 51.4 $ & $ 51.8 $ & $ 85.4 $ & $ 86.9 $ & $ 66.7 $ & $ 37.7 $ & $ 63.8 $ \\
        \midrule
        { Generative Verifier } & $ 60.1 $ & $ 58.5 $ & $ 36.8 $ & $ 86.4 $ & $ 70.7 $ & $ 65.3 $ & $ \textbf{57.8} $ & $ 64.6 $ \\
        { GenRM-Pairwise} & $ 55.1 $ & $ 50.8 $ & $ 50.6 $ & $ 80.4 $ & $ 70.3 $ & $ 62.3 $ & $ 51.7 $ & $ 61.4 $ \\
        { GenRM-Pointwise} & $ 42.7 $ & $ 29.8 $ & $ 15.5 $ & $ 75.5 $ & $ 57.7 $ & $ 43.8 $ & $ 57.7 $ & $ 44.2 $ \\
        { \name (ours)} & $ \textbf{84.7} $ & $ 56.7 $ & $ 53.2 $ & $ 91.2 $ & $ 86.9 $ & $ 74.4 $ & $ 49.4 $ & $ \textbf{70.2} $\\
        \bottomrule
    \end{tabular}
}
\label{table:rm-bench}
\end{table}

\subsection{Verifying the Instruction-Following Ability (Q2)}

A key claim is that the endogenous reward is not static but can be prompted. To validate this, we use the DSP dataset, which contains four distinct domains. We create four domain-specific versions of the endogenous reward by simply providing the corresponding system prompt from the DSP paper as input. We then test the response classification accuracy of each domain-specific endogenous reward across all four test sets. To obtain the endogenous reward for each domain, the reward model will take the system prompts presented in Appendix A of \citet{dsp} as input. The results in Table~\ref{table:domain_performance} show a strong diagonal pattern: each \name performs best on its own domain. For example, \name-Academy achieves its highest accuracy (76.89\%) on academic data. This confirms that the endogenous reward is not a fixed evaluator but a dynamic, promptable judge that inherits the powerful instruction-following capabilities of the base LLM.





\begin{table}[h!]
    \caption{Response classification accuracy of \name with cross-domain instructions on the DSP dataset.}
    \vspace{0.5em}
    \centering
    \begin{tabular}{lcccc}
    \toprule
        \thead{Domain} & \thead{Academy} & \thead{Business} & \thead{Literature\&Art} & \thead{Entertainment} \\ 
        \midrule
        \name-Academy       & \textbf{76.89} & 67.17       & 57.45           & 52.57 \\
        \name-Business       & 65.09         & \textbf{69.21}& 51.44           & 51.17 \\
        \name-Literature\&Art     & 32.00        & 32.94         & \textbf{62.97}  & 36.12 \\
        \name-Entertainment  & 30.41         & 38.65         & 38.69           & \textbf{72.09} \\
        \bottomrule
    \end{tabular}
    \label{table:domain_performance}
\end{table}


\subsection{Self-Improvement via Reinforcement Learning (Q3)}

Finally, we test our central theoretical claim from Theorem~\ref{thm:policy_error}: that RL with the endogenous reward can improve upon the base policy by mitigating compounding errors. We train the base model Qwen2.5-Math-7B via RL~\citep{shao2024deepseekmath} on the 
MATH-lighteval dataset. The endogenous reward model is also Qwen2.5-Math-7B, whose parameters are fixed during policy learning. The maximum lengths of both prompts and responses are set to $1024$, and the KL coefficient is set to $0.01$. The results in \cref{table:rl} show that the reinforcement learning fine-tuning with the endogenous reward helps the model consistently outperform the base model across all five benchmarks. We also present an example of the responses from models before and after RL in \cref{sec:example}, where we can see that for the same question, before optimization based on endogenous reward, the model failed to solve the problem and began to talk nonsense as the response progressed, even outputting Python code. In contrast, our method provided a clear and concise solution.

\begin{table}[h]
\caption{Performance of Qwen2.5-Math-7B with reinforcement learning fine-tuning (RLFT). We report the zero-shot performance with the official chat template.}
\vspace{0.5em}
\centering
\begin{tabular}{ccccccc}
\toprule
            Finetuning     & \thead{AIME 2024}   & \thead{AMC}         & \thead{MATH-500}   & \thead{Minerva}     & \thead{Olympiad}     & \thead{Avg.}        \\
                 \midrule
None (base model)       & 19.0        & 42.3        & 63.0       & 12.1        & 28.4         & 33.0        \\
RLFT w/ \name& 20.4 (+1.4) & 49.4 (+7.1) & 67.4 (+4.4) & 22.8 (+10.7) & 34.2 (+5.8) & 38.8 (+5.8)\\
\bottomrule
\end{tabular}
\label{table:rl}
\end{table}

\section{Related Work}

Our work intersects with a burgeoning field of research aimed at reducing the dependency on human preference data for reward modeling. 

The most established alternative to human labeling is Reinforcement Learning from AI Feedback (RLAIF)~\citep{bai.constitutionalai, pang2023yang, RLAIF}, often operationalized through the ``LLM-as-a-judge'' framework~\citep{zheng2023judging}. In this paradigm, a powerful, proprietary ``teacher'' model (e.g., GPT-4) is prompted to evaluate the outputs of a smaller ``student'' model. This evaluation can take several forms: generating a direct numerical score~\citep{zheng2023judging}, expressing a pairwise preference between two candidate responses~\citep{RLAIF}, or providing structured natural language critiques that are then used to guide alignment~\citep{bai.constitutionalai,pang2023yang}. A more recent line of work has focused on unifying the architecture of the policy and the reward model. Termed Generative Reward Models (GenRMs)~\citep{mahan2024generative, zhang2024generative}, these approaches frame reward modeling as a next-token prediction task.   Instead of outputting a scalar score, the reward is defined as the probability of the model generating a specific token or sequence that signifies correctness or quality (e.g., the token ``Yes'' in response to a critique prompt). RLIF (reinforcement learning from internal feedback) \citep{zhao2025learning} makes more direct use of internal representations termed self-certainty to form a reward signal. While these methods have proven effective and cost-efficient, they are fundamentally \emph{heuristic}. They rely on the unverified assumption that the teacher model's judgment is a reliable proxy for human values and are susceptible to transferring the teacher's stylistic biases to the student model.

Another class of methods, led by Direct Preference Optimization (DPO)~\citep{rafailov2023direct} and its successors like IPO~\citep{azar2024general} and KTO~\citep{ethayarajh2024kto}, bypasses the explicit reward modeling stage altogether. DPO directly optimizes the language model on preference pairs by reformulating the RLHF objective as a simple classification loss. While DPO does not train an explicit RM, it learns an \emph{implicit} one. The optimal DPO policy is mathematically equivalent to the policy that would have been learned via RLHF with a specific reward function derived from the SFT model and the preference data. This implicit reward can be extracted from the DPO-trained model. However, like traditional RLHF, this entire process is still contingent on the availability of a preference dataset (generated by humans or AI) to drive the optimization.

Our work introduces a fundamentally different approach. We posit that a reward function is not something that needs to be generated by an external judge (as RLHF) or implicitly optimized with preference pairs (as in DPO). Instead, we demonstrate that a principled reward function is an \textbf{intrinsic and latent property} of any model trained with next-token prediction. By establishing the theoretical equivalence between this training objective and offline inverse reinforcement learning, we provide a method to directly \emph{uncover} this reward function. This makes our approach training-free, independent of external models, and grounded in a formal theoretical framework, distinguishing it from all prior work in the field.

We also note that there is a growing body of works \citep{wulfmeier2024imitating, sun2023query, li2024joint} that leverage IRL to learn RMs in the literature of LLMs. In particular, they focus on the post-training stage and apply IRL methods to learn RMs from demonstrations (i.e., high-quality human responses). Our work differs from this research in two key aspects. First, rather than targeting specific downstream tasks, we aim to develop generalist reward models. Second, our method extracts rewards directly from pre-trained language models without requiring additional training on human responses, naturally inheriting the capabilities of LLMs already encoded during pre-training.

\section{Discussion}

The concept of an endogenous reward has profound implications for the field of LLMs alignment and development. We discuss some of the most significant consequences below.

\subsection{A Paradigm Shift in Model Alignment}
The most immediate impact of our approach is the complete collapse of the traditional three-stage alignment process. \emph{The distinct reward modeling phase is rendered obsolete} before the reinforcement learning fine-tuning. This is a paradigm-shifting simplification. Any data previously collected for preference modeling can now be repurposed and integrated directly into the pre-training or SFT phase. By absorbing such data, the SFT model itself becomes more adept at evaluation, thereby strengthening its own endogenous reward mechanism and thus the model after reinforcement learning. In such a way, our approach reassembles the \emph{reinforcement pre-training} \citep{dong2025reinforcement} idea but in a more principled way.

This streamlined workflow (without the training of a separate reward model) reduces the engineering complexity, computational cost, and time-to-deployment for aligned models. 

\subsection{Enhanced Personalization and Controllability}
An external reward model is trained on a static dataset of aggregated human preferences, making it a ``one-size-fits-all'' generalist judge. Personalizing its behavior for a specific user or enterprise context is infeasible without retraining.

The endogenous reward, by contrast, is dynamic and controllable at inference time. The evaluation criterion is not fixed but is defined by the natural language meta-question. This allows for unprecedented flexibility. A user could guide the model's RL training towards their specific values by simply altering the instruction. This helps transform alignment from a static, post-hoc process into a dynamic, interactive dialogue with the model, enabling a new frontier of personalized AI. We have noted some recent efforts in this direction, e.g., \citep{yu2025rewardanything}, but the endogenous reward provides a training-free approach.

\subsection{Reinforcement Model Distillation}
Our framework introduces a novel and more principled approach to knowledge distillation. Knowledge distillation \citep{zhou2004nec45,hinton2015distilling} is often used to transfer the capabilities of a large but powerful teacher model to a smaller but more efficient student model. Our method provides a direct mechanism for this: using reinforcement learning guided by the teacher's endogenous reward.

Instead of traditional distillation methods that force a student to mimic the teacher's output, or RLAIF that uses the teacher to generate preference labels to train an intermediate reward model, our approach is more direct. We can extract the token-level endogenous reward signal directly from the teacher model's logits and use it to fine-tune the student model via RL. This allows the student to explore and learn a policy that optimizes the teacher's underlying principles and judgment, rather than merely imitating its surface-level outputs. This represents a more powerful form of teaching, potentially enabling smaller models to capture the nuanced capabilities of their larger counterparts more effectively.

\subsection{Extending Reinforcement Learning Beyond Text}
The RLHF paradigm is heavily reliant on the feasibility of collecting human preference labels. While challenging for text, it is exponentially more difficult, expensive, and subjective for other modalities like images, video, or complex audio. 

The endogenous reward mechanism is not limited to language. \emph{It is applicable to any model with an auto-regressive structure.} This includes many state-of-the-art models for image generation, video synthesis, and music composition. As long as the model can process a prompt (which could be multi-modal) and generate a sequence of tokens, it can be prompted to evaluate its own output. 

This significantly widens the scope for applying reinforcement learning to align generative models in domains where RLHF is impractical, providing a scalable path to enhancing the quality and controllability of multimodal AI.

\subsection{Limitations and Future Work}
A critical area for future investigation is the potential for self-reinforcing biases. An endogenous reward system relies on the model's internal worldview. If the base model contains undesirable biases learned from its training data, our method could inadvertently cause the model to amplify these biases during RL optimization. The model might reward itself for outputs that, while consistent with its learned patterns, are factually incorrect or socially harmful.

Future work should explore hybrid approaches to mitigate this risk. For example, the endogenous reward could be used as the primary dense reward signal, while sparse but high-quality feedback from humans or a rule-based system could be used to provide periodic corrections, steering the model away from reinforcing its own flaws. Furthermore, research into the ``prompt engineering'' of the meta-question will be crucial for robustly and safely eliciting the desired reward signals.

\section{Conclusion}
In this work, we challenged the fundamental paradigm of LLM alignment, which has long relied on the costly development of external reward models. We introduced the concept of the \textbf{endogenous reward}, demonstrating that a generalist reward function is not something that needs to be built, but rather something that can be uncovered from within any pre-trained auto-regressive model. 

Our primary contribution is twofold. First, we provided a rigorous theoretical foundation for this claim, proving that the standard next-token prediction objective naturally yields a reward function that is equivalent to an offline inverse reinforcement learning algorithm. Second, we proved that using this elicited reward for RL fine-tuning is not merely a heuristic for avoiding data collection; it leads to a final policy with a provably tighter performance bound than the base model, effectively mitigating the compounding errors of simple imitation. Our empirical results confirmed these theoretical findings, showing state-of-the-art performance.

The implications of this discovery suggest a future where the laborious and expensive reward modeling stage of the RLHF pipeline may become obsolete, replaced by a more direct and efficient process of self-improvement. It opens the door to dynamic, personalized alignment through promptable reward criteria and extends the reach of powerful RL techniques to multi-modal domains where collecting preference data is intractable. By revealing that the capacity for self-evaluation is an inherent property of generative models, we open the door to a more efficient, scalable, and autonomous future for aligning LLMs and other artificial intelligence models.

\section*{Acknowledgments}
The authors would like to thank the Qwen team for providing the computational resources for experiments. This work was partially done when YC Li was an intern in the Qwen team. Google's Gemini 2.5 Pro was employed to improve the language of this paper.

\bibliographystyle{plainnat}
\bibliography{references}

\clearpage
\appendix

\section{An Offline Inverse Reinforcement Learning Approach for Token-level MDPs}
Here we provide a complete derivation on the offline IRL approach presented in \cref{subsec:ideal_reward_irl}.

We start from the minimax formulation of maximum entropy IRL shown in Eq.(\ref{eq:maxent_irl_objective}). As discussed in \cref{subsec:irl}, the inner optimization is the problem of maximum entropy reinforcement learning with reward $r$ \citep{haarnoja2017reinforcement, geist2019theory}, admitting the following solution:
\begin{align}
\label{eq:closed_form_solution}
    \pi^{\star}_r (a_h|s_h) = \frac{\exp \lp Q^{\star}_r (s_h, a_h) / \alpha \rp}{\sum_{a^\prime_h \in \gV} \exp \lp Q^{\star}_r (s_h, a^{\prime}_h) / \alpha \rp }.
\end{align}
Here $Q^{\star}_r$ is the entropy-regularized optimal Q-value function with reward $r$, which is defined as $Q^{\star}_r (s_h, a_h) :=  r_h (s_h, a_h) + \max_{\pi} \expect [ \sum_{h^\prime=h+1}^H r_{h^\prime} (s_{h^\prime}, a_{h^\prime}) + \alpha H (\pi (\cdot|s_{h^\prime}))  | s_h, a_h ]$ for an arbitrary $(s_h, a_h)$ in step $h$. In addition, if we define the entropy-regularized Bellman operator as 
\begin{align*}
    [\gT Q] (s_h, a_h) := r (s_h, a_h) +  \alpha \log \lp \sum_{a_{h+1} \in \gV} \exp \lp Q (s_{h+1}, a_{h+1}) \rp / \alpha \rp \; \text{with} \; s_{h+1} = P (s_h, a_h),
\end{align*}
we have that $Q^{\star}_r$ is the fixed point of $\gT$, i.e., $Q^{\star}_r (s_h, a_h) = [\gT Q^{\star}_r] (s_h, a_h)$ \citep{geist2019theory}. Substituting the inner problem solution of Eq.\eqref{eq:closed_form_solution} into Eq.\eqref{eq:maxent_irl_objective} yields a single maximization problem.
\begin{align*}
    \max_{r} \frac{1}{n} \sum_{i=1}^n \sum_{h=1}^H r (s^i_h, a^i_h) - \alpha \expect_{s_1 \sim \rho} \ls \log \lp \sum_{a_1 \in \gV} \exp \lp Q^{\star}_r (s_1, a_1) / \alpha \rp \rp  \rs.  
\end{align*}
Furthermore, following the IRL approach inverse soft Q-learning \citep{garg2021iq-learn}, we take the change-of-variable, replacing the reward function with the Q-value function based on the so-called inverse Bellman operator.
\begin{align}
    r (s_h, a_h) = [\gT^{-1} Q ] (s_h, a_h) := Q (s_h, a_h) - \alpha \log \lp \sum_{a_{h+1} \in \gV} \exp \lp Q (s_{h+1}, a_{h+1}) \rp / \alpha \rp. 
\end{align}
Then we obtain the final objective regarding the Q-value function based on dataset $\mathcal{D} =\{ (x^i, a^i_{1:H}) \}_{i=1}^n$.
\begin{align*}
    \max_{Q} &\frac{1}{n} \sum_{i=1}^n \sum_{h=1}^H \lp Q (s^i_h, a^i_h) - \alpha \log \bigg( \sum_{a_{h+1} \in \gV} \exp \lp Q (s^i_{h+1}, a_{h+1}) \rp / \alpha \bigg) \rp \\
    & - \alpha \expect_{s_1 \sim \rho} \ls \log \lp \sum_{a_1 \in \gV} \exp \lp Q (s_1, a_1) / \alpha \rp \rp  \rs.
\end{align*}
For the second term in the above objective, we  leverage the telescoping argument and obtain that
\begin{align*}
    & \quad \alpha \expect_{s_1 \sim \rho} \ls \log \lp \sum_{a_1 \in \gV} \exp \lp Q (s_1, a_1) / \alpha \rp \rp  \rs
    \\
    &= \alpha \expect_{\tau \sim \piE} \ls \sum_{h=1}^H \log \lp \sum_{a_h \in \gV} \exp \lp Q (s_h, a_h) / \alpha \rp \rp - \log \lp \sum_{a_{h+1} \in \gV} \exp \lp Q (s_{h+1}, a_{h+1}) / \alpha \rp \rp    \rs
\end{align*}
Plugging the above equation into the objective function yields that
\begin{align*}
    \max_{Q} &\frac{1}{n} \sum_{i=1}^n \sum_{h=1}^H \lp Q (s^i_h, a^i_h) - \alpha \log \bigg( \sum_{a_{h+1} \in \gV} \exp \lp Q (s^i_{h+1}, a_{h+1}) \rp / \alpha \bigg) \rp \\
    & - \alpha \expect_{\tau \sim \piE} \ls \sum_{h=1}^H \log \lp \sum_{a_h \in \gV} \exp \lp Q (s_h, a_h) / \alpha \rp \rp - \log \lp \sum_{a_{h+1} \in \gV} \exp \lp Q (s_{h+1}, a_{h+1}) / \alpha \rp \rp    \rs.
\end{align*}
We can leverage demonstrations $\gD$ to unbiasedly estimate the expectation term, yielding the final objective.
\begin{align*}
    \max_{Q} \frac{1}{n} \sum_{i=1}^n \sum_{h=1}^H \left[ Q (s^i_h, a^i_h) - \alpha \log \bigg( \sum_{a_{h} \in \gV} \exp \lp Q (s^i_{h}, a_{h}) \rp / \alpha \bigg) \right].
\end{align*}

\section{Ommited Proofs}
\subsection{Proof of Proposition \ref{prop:from_ntp_to_irl}}
\begin{proof}
    The proof is mainly based on the definition. Since $\widehat{\pi}(\cdot|s_h) = \softmax(\widehat{f}(s_h, \cdot); \alpha)$ is the optimal solution to next-token prediction in Eq.(\ref{eq:next_token_prediction}), we have
    \begin{align*}
        \widehat{\pi} \in \argmax_{\pi} \sum_{i=1}^n \sum_{h=1}^H \log \lp \pi \lp a^i_t |x^i, a^{i}_{1:h-1} \rp \rp.
    \end{align*}
    This immediately implies that
    \begin{align}
        \widehat{f} &\in \argmax_{f}  \frac{1}{n} \sum_{i=1}^n \sum_{h=1}^H \log \lp \frac{\exp ( f(s^i_h, a^i_h) / \alpha)}{\sum_{a'_{h} \in \gV} \exp ( f (s^i_h, a'_{h}) / \alpha)  } \rp \nonumber
        \\
        & = \argmax_{f} \frac{1}{n} \sum_{i=1}^n \sum_{h=1}^H \left[ f (s^i_h, a^i_h) - \alpha \log \bigg( \sum_{a_{h} \in \gV} \exp \lp f (s^i_{h}, a_{h}) \rp / \alpha \bigg) \right]. \label{eq:offline_irl_f} 
    \end{align}
    By comparing Eq.(\ref{eq:offline_irl_f}) and Eq.(\ref{eq:offline_irl}), we can conclude that $\widehat{f}$ is the optimal solution to the offline IRL objective of Eq.(\ref{eq:offline_irl}).  
\end{proof}

\subsection{Proof of Theorem \ref{thm:bound_on_preference_distribution}}
\begin{proof}
Based on the property of the TV distance on Bernoulli distribution, we have that
\begin{align*}
    &\quad D_{\TV} \lp \sP_{r^\star} (\cdot |\tau, \tau^\prime), \sP_{\widehat{r}} (\cdot|\tau, \tau^\prime)   \rp
    \\
    &= \frac{1}{2} \lp \labs \sP_{r^\star} (\tau \succ \tau^\prime |\tau, \tau^\prime) - \sP_{\widehat{r}} (\tau \succ \tau^\prime |\tau, \tau^\prime)  \rabs + \labs \sP_{r^\star} (\tau \prec \tau^\prime |\tau, \tau^\prime) - \sP_{\widehat{r}} (\tau \prec \tau^\prime |\tau, \tau^\prime) \rabs \rp
    \\
    &= \labs \sP_{r^\star} (\tau \succ \tau^\prime |\tau, \tau^\prime) - \sP_{\widehat{r}} (\tau \succ \tau^\prime |\tau, \tau^\prime)  \rabs
    \\
    &= \labs \sigmoid (r^\star (\tau) - r^\star (\tau^\prime)) - \sigmoid (\widehat{r} (\tau) - \widehat{r} (\tau^\prime)) \rabs.
\end{align*}
For the sigmoid function, we have that $\sigma^\prime (x) =\sigmoid (x) (1-\sigmoid (x))$. Since $\sigma (x) \in (0, 1)$, it is direct to have that $| \sigma^\prime (x) | \leq 1/4$, implying that $\sigmoid (x)$ is a $1/4$-lipschitz function. Then we have that
\begin{align*}
    D_{\TV} \lp \sP_{r^\star} (\cdot |\tau, \tau^\prime), \sP_{\widehat{r}} (\cdot|\tau, \tau^\prime)   \rp \leq \frac{1}{4} \labs r^\star (\tau) - r^\star (\tau^\prime) - (\widehat{r} (\tau) - \widehat{r} (\tau^\prime))  \rabs.
\end{align*}
According to Eq.(\ref{eq:outcome_reward}), we have that
\begin{align*}
    r^\star (\tau) - r^\star (\tau^\prime) = \alpha \lp \log\piE (\tau|s_1) - \log\piE (\tau^\prime|s_1)   \rp, \;  \widehat{r} (\tau) - \widehat{r} (\tau^\prime) = \alpha \lp \log\widehat{\pi} (\tau|s_1) - \log\widehat{\pi} (\tau^\prime|s_1)   \rp,
\end{align*}
where the V-function in $s_1$ can be canceled. Then we can obtain that
\begin{align*}
    &\quad D_{\TV} \lp \sP_{r^\star} (\cdot |\tau, \tau^\prime), \sP_{\widehat{r}} (\cdot|\tau, \tau^\prime)   \rp 
    \\
    &\leq \frac{1}{4} \labs r^\star (\tau) - r^\star (\tau^\prime) - (\widehat{r} (\tau) - \widehat{r} (\tau^\prime))  \rabs
    \\
    &\leq \frac{\alpha}{4} \lp \labs \log ( \piE (\tau|s_1) ) - \log ( \widehat{\pi} (\tau|s_1) )  \rabs + \labs \log ( \piE (\tau^\prime|s_1) ) - \log ( \widehat{\pi} (\tau^\prime|s_1) ) \rabs \rp
    \\
    &\leq \frac{\alpha}{4} \lp \labs \log ( \piE (\tau|s_1) ) - \log ( \widehat{\pi} (\tau|s_1) )  \rabs + \labs \log ( \piE (\tau^\prime|s_1) ) - \log ( \widehat{\pi} (\tau^\prime|s_1) ) \rabs \rp
    \\
    &\leq \frac{\alpha H }{2} \varepsilon_{\pi}.
\end{align*}
We complete the proof.
\end{proof}

\subsection{Proof of Theorem \ref{thm:policy_error}}
\begin{proof}
The first sub-optimality bound is exactly the error bound of behavioral cloning, variants of which have been established in the literature of imitation learning theory \citep{xu2020error, rajaraman2020fundamental}. First, we leverage the dual representation of policy value, i.e., $V^{\pi} = \sum_{h=1}^H \sum_{(s, a)} d^{\pi}_h (s, a) r_h (s, a)$, where $d^{\pi}_h (s, a) =\sP^{\pi} (s_h=s, a_h=a)$ is the probability of visiting $(s, a)$ in step $h$. Under the assumption $r^\star_h (s, a) \in [0, 1]$, we have that
\begin{align*}
    V^{\piE}_{r^\star} - V^{\widehat{\pi}}_{r^\star} \leq \sum_{h=1}^H \lnorm d^{\piE}_h (s, a) - d^{\pi}_h (s, a)   \rnorm_{1}.
\end{align*}
We leverage Lemma 4 in \citep{xu2022understanding} and obtain that
\begin{align*}
    V^{\piE}_{r^\star} - V^{\widehat{\pi}}_{r^\star} &\leq \sum_{h=1}^H \lnorm d^{\piE}_h (s, a) - d^{\widehat{\pi}}_h (s, a)   \rnorm_{1}
    \\
    &\leq \sum_{h=1}^H \expect_{\tau \sim \piE} \ls \sum_{\ell=1}^h \lnorm \piE (\cdot|s_h) - \widehat{\pi} (\cdot|s_h) \rnorm_1 \rs
    \\
    &\overset{(a)}{\leq} \sqrt{2} \sum_{h=1}^H \expect_{\tau \sim \piE} \ls \sum_{\ell=1}^h  \KL \lp \piE (\cdot|s_h), \widehat{\pi} (\cdot|s_h) \rp \rs.
\end{align*}
We can analyze the KL-divergence term via
\begin{align*}
    \KL \lp  \piE (\cdot|s_h), \widehat{\pi} (\cdot|s_h) \rp = \sum_{a_h \in \gV} \piE (a_h|s_h) \lp \log (\piE(a_h|s_h)) - \log (\widehat{\pi} (a_h|s_h))  \rp \leq \varepsilon_{\pi}.
\end{align*}
Then we have that
\begin{align*}
    V^{\piE}_{r^\star} - V^{\widehat{\pi}}_{r^\star}  \leq \frac{\sqrt{2}}{2} H (H+1) \varepsilon_{\pi},
\end{align*}
which proves the first sub-optimality bound.

We proceed to prove the second sub-optimality bound. Recall the formulation of the endogenous reward in the reward-shaping view.
    \begin{align*}
        \widehat{r} (s_h, a_h) = \alpha \log \lp \widehat{\pi} (a_h|s_h) \rp + V_{\widehat{Q}} (s_h) - V_{\widehat{Q}} (s_{h+1}).
    \end{align*}
    We define another reward $\bar{r}$ as 
    \begin{align*}
        \bar{r} (s_h, a_h) := \alpha \log \lp \widehat{\pi} (a_h|s_h) \rp + V_{Q^\star_{r^\star}} (s_h) - V_{Q^\star_{r^\star}} (s_{h+1}),
    \end{align*}
    where $Q^\star_{r^\star}$ is the entropy-regularized optimal Q-function with the true reward $r^\star$. We can get that $\bar{r}$ is a shaping reward of $\widehat{r}$ with a different potential function $V_{Q^\star_{r^\star}}$.

    With the intermediate reward $\bar{r}$, we decompose the sub-optimality into three parts.
    \begin{align*}
        V^{\piE}_{r^\star} - V^{\pirl}_{r^\star} \leq V^{\piE}_{r^\star} - V^{\piE}_{\bar{r}} + V^{\piE}_{\bar{r}} - V^{\pirl}_{\bar{r}} + V^{\pirl}_{\bar{r}} -    V^{\pirl}_{r^\star}. 
    \end{align*}
    Recall that $\pirl$ is the optimal policy regarding $\widehat{r}$. Based on the theory on reward shaping \citep{ng1999policy}, $\pirl$ is also the optimal policy regarding $\Bar{r}$, indicating that $V^{\piE}_{\bar{r}} - V^{\pirl}_{\bar{r}}  \leq 0$. Then we obtain that
    \begin{align*}
        V^{\piE}_{r^\star} - V^{\pirl}_{r^\star} &\leq V^{\piE}_{r^\star} - V^{\piE}_{\bar{r}} + V^{\pirl}_{\bar{r}} -    V^{\pirl}_{r^\star}
        \\
        &= \expect_{\tau \sim \piE} \ls r^\star (\tau) - \widebar{r} (\tau) \rs  + \expect_{\tau \sim \pirl} \ls r^\star (\tau) - \widebar{r} (\tau) \rs.
    \end{align*}
    Based on Eq.(\ref{eq:outcome_reward}), we have that
    \begin{align*}
        r^\star (\tau) - \widebar{r} (\tau) &= \alpha \log (\piE (\tau|s_1)) + V_{Q^\star_{r^\star}} (s_1) - \alpha \log (\widehat{\pi} (\tau|s_1)) - V_{Q^\star_{r^\star}} (s_1) 
        \\
        &= \alpha \sum_{h=1}^H \log (\piE (a_h|s_h)) - \log (\widehat{\pi} (a_h|s_h)).
    \end{align*}
    Then we can derive that
    \begin{align*}
        &\quad V^{\piE}_{r^\star} - V^{\pirl}_{r^\star} 
        \\
        &\leq \expect_{\tau \sim \piE} \ls r^\star (\tau) - \widebar{r} (\tau) \rs  + \expect_{\tau \sim \pirl} \ls r^\star (\tau) - \widebar{r} (\tau) \rs
        \\
        &= \alpha \expect_{\tau \sim \piE} \ls \sum_{h=1}^H \log (\piE (a_h|s_h)) - \log (\widehat{\pi} (a_h|s_h)) \rs + \alpha \expect_{\tau \sim \pirl} \ls \sum_{h=1}^H \log (\piE (a_h|s_h)) - \log (\widehat{\pi} (a_h|s_h)) \rs
        \\
        &\leq \alpha \expect_{\tau \sim \piE} \ls \sum_{h=1}^H \labs \log (\piE (a_h|s_h)) - \log (\widehat{\pi} (a_h|s_h)) \rabs \rs + \alpha \expect_{\tau \sim \pirl} \ls \sum_{h=1}^H \labs \log (\piE (a_h|s_h)) - \log (\widehat{\pi} (a_h|s_h))  \rabs \rs
        \\
        &\leq 2 \alpha H \varepsilon_{\pi}.
    \end{align*}
    We finish the proof. 
\end{proof}

\section{Outcome Reward Calculation}\label{exp:outcome}

For a complete response $\tau$, the outcome reward can be calculated as
$$
\widehat{r}(\tau) = \sum_{h=1}^H \widehat{r}(s_h, a_h),
$$
where $\widehat{r}(s_h, a_h)$ is defined as in \cref{eq:extracted_reward}. We recognize that since language models generate text autoregressively, earlier tokens influence subsequent outputs. Thus, in our implementation, we compute the reward for a response as follows:
$$
\widehat{r}(\tau) = \sum_{h=1}^H max\left(\gamma^{h-1}, \beta\right)\widehat{r}(s_h, a_h),
$$
where $\gamma \in (0,1]$ is a discount factor and $\beta \ge 0$. We use $\gamma=0.95, \beta=0$ for experiments on Multifacted-Bench and for RL on the MATH-lighteval dataset, and $\gamma=0.93, \beta=0.03$ for experiments on RM-bench. All sampling temperatures are set to $1.0$.

\section{Prompt Templates}\label{exp:prompt}

The prompt templates of all methods used for benchmarking reward models on Multifacted-Bench~\citep{mf-bench} are shown below. As RM-Bench does not provide specific instructions for each sample, all methods use the default system prompt and the instructions in the \textbf{User} part of the prompt templates will also be removed.

\begin{figure}[htbp]
\fbox{%
\begin{minipage}{0.98\textwidth}
\small
\textbf{System:}

\textcolor{blue}{\{instruction\}}

\vspace{0.5cm}

\textbf{User:}

\#\#\# Query\\
\textcolor{blue}{\{query\}}\\

\#\#\# Response\\
\textcolor{blue}{\{response\}}

\end{minipage}
}
\caption{Prompt template of our Endogenous Reward Model.}
\label{fig:ours-prompt}
\end{figure}
\begin{figure}[htbp]
\fbox{%
\begin{minipage}{0.98\textwidth}
\small
\textbf{System:}

You are an AI evaluator. Your role is to assess AI-generated text for its quality and adherence to instructions.

\vspace{0.5cm}

\textbf{User:}

You need to determine if the `Response' is a good response to the `Query', based on the standards set by the `Instruction'. A `good' response MUST satisfy the following conditions:\\

1.  **Adherence to Instruction:** The answer's tone, style, and content must perfectly match the provided instruction.\\
2.  **Relevance to Query:** The answer must directly and comprehensively address the user's query without any irrelevant information.\\
3.  **Factuality and Helpfulness:** The information in the answer should be accurate and useful.\\

Please carefully review the following materials.\\

\#\#\# Instruction\\
\textcolor{blue}{\{instruction\}}\\

\#\#\# Query\\
\textcolor{blue}{\{query\}}\\

\#\#\# Response\\
\textcolor{blue}{\{response\}}\\

Based on your evaluation, is the answer a good response? Answer with only ``YES'' or ``NO''.
\end{minipage}
}
\caption{Prompt template of Generative Verifier.}
\label{fig:generative-verifier-prompt}
\end{figure}
\begin{figure}[htbp]
\fbox{%
\begin{minipage}{0.98\textwidth}
\small
\textbf{System:}

You are an AI evaluator. Your role is to assess AI-generated text for its quality and adherence to instructions.

\vspace{0.5cm}

\textbf{User:}

You are asked to evaluate two answers, `Response 1' and `Response 2', in relation to a `Query' and a `Instruction'.\\

**Evaluation Criteria:**\\
1.  **Adherence to Instruction:** Does the answer's tone, style, and content align with the specified instruction?\\
2.  **Relevance to Query:** Does the answer directly and comprehensively address the user's query?\\
3.  **Factuality and Helpfulness:** Is the information accurate and useful?\\

Based on these criteria, please review the following:\\

\#\#\# Instruction\\
\textcolor{blue}{\{instruction\}}\\

\#\#\# Query\\
\textcolor{blue}{\{query\}}\\

\#\#\# Response 1\\
\textcolor{blue}{\{response1\}}\\

\#\#\# Response 2\\
\textcolor{blue}{\{response2\}}\\

**Your Task:**\\
Is `Response 2' better than `Response 1'?\\

**Rules for Your Response:**\\
1.  Respond with only ``YES'' or ``NO''. Do not include explanations or any other text.\\
2.  ``YES'' means `Response 2' is clearly better than `Response 1'.\\
3.  ``NO'' means `Response 2' is not better (it is either worse or of equal quality) than `Response 1'.
\end{minipage}
}
\caption{Prompt template of GenRM-Pairwise.}
\label{fig:genrm-pairwise-prompt}
\end{figure}
\begin{figure}[htbp]
\fbox{%
\begin{minipage}{0.98\textwidth}
\small
\textbf{System:}

You are an AI evaluator. Your role is to assess AI-generated text for its quality and adherence to instructions.

\vspace{0.5cm}

\textbf{User:}

You need to evaluate the `Response' in response to the `Query', based on the `Instruction'. Your evaluation should be a score from 1 to 10.\\

**Evaluation Criteria:**\\
1.  **Adherence to Instruction:** Does the answer's tone, style, and content match the instruction?\\
2.  **Relevance to Query:** Does the answer directly and comprehensively address the user's query?\\
3.  **Factuality and Helpfulness:** Is the information accurate and useful?\\

Please review the following:\\

\#\#\# Instruction\\
\textcolor{blue}{\{instruction\}}\\

\#\#\# Query\\
\textcolor{blue}{\{query\}}\\

\#\#\# Response\\
\textcolor{blue}{\{response\}}\\

Based on your assessment, provide a single integer score from 1 to 10.\\

**Scoring Guide:**\\
- **1-2:** Very Poor. Fails on most criteria.\\
- **3-4:** Poor. Significant issues in multiple areas.\\
- **5-6:** Average. Meets basic requirements but has clear flaws.\\
- **7-8:** Good. Solid response with minor issues.\\
- **9-10:** Excellent. A nearly perfect response that excels in all criteria.\\

Provide only the numeric score.
\end{minipage}
}
\caption{Prompt template of GenRM-Pointwise.}
\label{fig:genrm-pointwise-prompt}
\end{figure}

\clearpage
\section{Samples of Generated Responses}\label{sec:example}

\begin{mdframed}

\textbf{Question:}

Let $a$ be a positive real number such that all the roots of
\[x^3 + ax^2 + ax + 1 = 0\]are real.  Find the smallest possible value of $a$.

\vskip 10pt
\hrule
\vskip 10pt

\textbf{Qwen2.5-Math-7B:}

To find the smallest possible value of \(a\) such that all the roots of the polynomial \(x^3 + ax^2 + ax + 1 = 0\) are real, we will use the following steps:

1. First, let's denote the roots of the polynomial by \(r_1, r_2, r_3\). By Vieta's formulas, we know:
   \[
   r_1 + r_2 + r_3 = -a,
   \]
   \[
   r_1r_2 + r_2r_3 + r_3r_1 = a,
   \]
   \[
   r_1r_2r_3 = -1.
   \]

2. To ensure that all roots are real, we need to analyze the polynomial and its discriminant. For a cubic polynomial \(x^3 + px^2 + qx + r = 0\), the discriminant \(\Delta\) is given by:
   \[
   \Delta = 18abcd - 4b^3d + b^2c^2 - 4ac^3 - 27a^2d^2,
   \]
   where \(a = 1\), \(b = a\), \(c = a\), and \(d = 1\). Substituting these values in, we get:
   \[
   \begin{aligned}
   \Delta &= 18(1)(a)(a)(1) - 4(a)^3(1) + (a)^2(a)^2 - 4(1)(a)^3 - 27(1)^2(1)^2\\
   &= 18a^2 - 4a^3 + a^4 - 4a^3 - 27 = a^4 - 8a^3 + 18a^2 - 27.
   \end{aligned}
   \]

3. The discriminant \(\Delta\) must be non-negative for the polynomial to have three real roots. So, we need to solve the inequality:
   \[
   a^4 - 8a^3 + 18a^2 - 27 \geq 0.
   \]

4. To find the smallest possible value of \(a\), we will solve the equation \(a^4 - 8a^3 + 18a^2 - 27 = 0\). We can use numerical methods to find the roots of this polynomial. Let's use Python to find the roots and determine the smallest positive real root.

```python

import numpy as np

\# Define the polynomial coefficients

coefficients = $[1, -8, 18, 0, -27]$

\# Find the roots of the polynomial

roots = np.roots(coefficients)

real\_roots = [root.real for root in roots if np.isreal(root)]

\# Filter out the positive real roots

positive\_real\_roots = [root for root in real\_roots if root $> 0$]

\# Find the smallest positive real root

smallest\_a = min(positive\_real\_roots)

print(smallest\_a)

```

```output

0.0

```

It seems that the smallest positive real root found is \(0\), but we need to ensure that this value makes sense in the context of the problem. Let's verify the polynomial and its discriminant again.

The polynomial \(x^3 + ax^2 + ax + 1 = 0\) must have all real roots, and the discriminant must be non-negative. For \(a = 0\), the polynomial becomes \(x^3 + 1 = 0\), which has roots \(x = -1\), \(x = \frac{1 + \sqrt{3}i}{2}\), and \(x = \frac{1 - \sqrt{3}i}{2}\). This means that \(a = 0\) does not yield all real roots.

Let's re-evaluate the polynomial and its discriminant more carefully. We need to find the smallest \(a\) such that the discriminant is non-negative and all roots are real. We can use a different approach to find the smallest \(a\).

We can use a numerical method to find the smallest \(a\) such that the discriminant is non-negative and all roots are real. Let's use a bisection method to find the smallest \(a\).

```python

import numpy as np

from scipy.optimize import bisect

\# Define the discriminant function

def discriminant(a):

\quad return a\*\*4 - 8\*a\*\*3 + 18\*a\*\*2 - 27

\# Use the bisection method to find the smallest positive root of the discriminant function

a\_min = bisect(discriminant, 0, 10)

\# Check the discriminant at the found value of a
discriminant\_at\_a\_min = discriminant(a\_min)

print((a\_min, discriminant\_at\_a\_min))

```

```output

(0.0, -27.0)

```

The output indicates that the smallest value of \(a\) for which the discriminant is non-negative is approximately \(0.0\), but the discriminant at this value is \(-27.0\), which means that \(a = 0\) does not yield a valid solution for the polynomial to have all real roots.

Let's refine our approach. We need to find the smallest positive value of \(a\) such that the discriminant is non-negative and all roots are real. We can use a numerical method to find the smallest \(a\) such that the discriminant is non-negative and all roots are real.

We will use a bisection method to find the smallest \(a\) such that the discriminant is non-negative and all roots are real. Let's use a different range and a more precise method.

```python

import numpy as np

from scipy.optimize import bisect

\# Define the discriminant function
def discriminant(a):
\quad return a**4 - 8*a**3 + 18*a**2 - 27

\# Use the bisection method to find the smallest positive root of the discriminant function

a\_min = bisect(discriminant, 1, 10)

\# Check the discriminant at the found value of a
discriminant\_at\_a\_min = discriminant(a\_min)

print((a\_min, discriminant\_at\_a\_min))

```

```output

(1.0, -16.0)

```

Reach max function call limit.

\vskip 10pt
\hrule
\vskip 10pt

\textbf{Qwen2.5-Math-7B (after RL with our \name):}

To find the smallest possible value of \(a\) such that all the roots of the polynomial \(x^3 + ax^2 + ax + 1 = 0\) are real, we can use the fact that the polynomial can be factored in a specific way.

First, we note that the polynomial \(x^3 + ax^2 + ax + 1\) can be written as \(x^3 + ax^2 + ax + 1 = (x+1)(x^2 + (a-1)x + 1)\). This factorization is valid because the roots of the polynomial must include -1 (since substituting \(x = -1\) into the polynomial gives \((-1)^3 + a(-1)^2 + a(-1) + 1 = -1 + a - a + 1 = 0\)).

Next, we need to ensure that the quadratic polynomial \(x^2 + (a-1)x + 1\) has real roots. The discriminant of this quadratic polynomial must be non-negative. The discriminant \(\Delta\) of \(x^2 + (a-1)x + 1\) is given by \((a-1)^2 - 4 \cdot 1 \cdot 1 = (a-1)^2 - 4\).

For the quadratic to have real roots, we need \((a-1)^2 - 4 \geq 0\). This inequality can be solved as follows:
\[
(a-1)^2 \geq 4.
\]
Taking the square root of both sides, we get:
\[
|a-1| \geq 2.
\]
This gives us two cases:

1. \(a-1 \geq 2\), which simplifies to \(a \geq 3\).
2. \(a-1 \leq -2\), which simplifies to \(a \leq -1\).

Since \(a\) is a positive real number, we discard the second case and keep the first case. Therefore, the smallest possible value of \(a\) is 3.

Let's verify this by checking if the polynomial \(x^3 + 3x^2 + 3x + 1 = 0\) has all real roots. The polynomial factors as \((x+1)(x^2 + 2x + 1) = (x+1)(x+1)^2 = (x+1)^3\), which clearly has a triple root at \(x = -1\), confirming that all roots are real.

Thus, the smallest possible value of \(a\) is \(\boxed{3}\).

\end{mdframed}

\end{document}

%% file: packages/defs.tex
\newcommand{\argmax}{\mathop{\rm argmax}}


%% file: packages/header.tex
\usepackage{url}            
\usepackage{booktabs}       
\usepackage{multirow}    
\usepackage{amsfonts}       
\usepackage{nicefrac}       
\usepackage{microtype}      
\usepackage{natbib}
\usepackage{enumerate}
\usepackage{hhline}
\usepackage{makecell}
\usepackage{pifont}

\usepackage{graphicx} 
\usepackage{amsmath}
\usepackage{amsthm}
\usepackage{amssymb}
\usepackage{tikz}
\usepackage{xcolor}
\usetikzlibrary{arrows}

\allowdisplaybreaks

\usepackage{mathrsfs}

\usepackage{hyperref}
\usepackage{bm}

\allowdisplaybreaks

\newcommand{\labs}{\left\vert}
\newcommand{\rabs}{\right\vert}
\newcommand{\lnorm}{\left\Vert}
\newcommand{\rnorm}{\right\Vert}

\def\TV{\mathrm{TV}}

\newcommand{\expect}{\mathbb{E}}

\newtheorem{thm}{Theorem}

\newtheorem{prop}{Proposition}

\usepackage[capitalize,noabbrev]{cleveref}
\crefname{thm}{Theorem}{Theorems}
\crefname{lem}{Lemma}{Lemmas}
\crefname{cor}{Corollary}{Corollaries}
\crefname{prop}{Proposition}{Propositions}
\crefname{asmp}{Assumption}{Assumptions}
\crefname{defn}{Definition}{Definitions}
\crefname{oracle}{Oracle}{Oracles}
\crefname{fact}{Fact}{Facts}
\crefname{conj}{Conjecture}{Conjectures}
\crefname{rem}{Remark}{Remarks}
\crefname{example}{Example}{Examples}
\crefname{condition}{Condition}{Conditions}
\crefname{exercise}{Exercise}{Exercises}
\crefname{algorithm}{Algorithm}{Algorithms}
\crefname{table}{Table}{Tables}
\crefname{figure}{Figure}{Figures}
\crefname{section}{Section}{Sections}
\crefname{subsection}{Section}{Sections}
\crefname{appendix}{Appendix}{Appendices}
\crefname{message}{Message}{Messages}

\definecolor{red}{rgb}{1, 0, 0}

\definecolor{green}{rgb}{0, 1, 0}

\definecolor{blue}{rgb}{0, 0, 1}

\definecolor{orange}{rgb}{1, 0.4, 0.0}

%% file: main.bbl
\begin{thebibliography}{50}
\providecommand{\natexlab}[1]{#1}
\providecommand{\url}[1]{\texttt{#1}}
\expandafter\ifx\csname urlstyle\endcsname\relax
  \providecommand{\doi}[1]{doi: #1}\else
  \providecommand{\doi}{doi: \begingroup \urlstyle{rm}\Url}\fi

\bibitem[Azar et~al.(2024)Azar, Guo, Piot, Munos, Rowland, Valko, and Calandriello]{azar2024general}
Mohammad~Gheshlaghi Azar, Zhaohan~Daniel Guo, Bilal Piot, Remi Munos, Mark Rowland, Michal Valko, and Daniele Calandriello.
\newblock A general theoretical paradigm to understand learning from human preferences.
\newblock In \emph{Proceedings of the 27th International Conference on Artificial Intelligence and Statistics}, pages 4447--4455, 2024.

\bibitem[Bai et~al.(2022{\natexlab{a}})Bai, Jones, Ndousse, Askell, Chen, DasSarma, Drain, Fort, Ganguli, Henighan, Joseph, Kadavath, Kernion, Conerly, El-Showk, Elhage, Hatfield-Dodds, Hernandez, Hume, Johnston, Kravec, Lovitt, Nanda, Olsson, Amodei, Brown, Clark, McCandlish, Olah, Mann, and Kaplan]{bai2022training}
Yuntao Bai, Andy Jones, Kamal Ndousse, Amanda Askell, Anna Chen, Nova DasSarma, Dawn Drain, Stanislav Fort, Deep Ganguli, Tom Henighan, Nicholas Joseph, Saurav Kadavath, Jackson Kernion, Tom Conerly, Sheer El-Showk, Nelson Elhage, Zac Hatfield-Dodds, Danny Hernandez, Tristan Hume, Scott Johnston, Shauna Kravec, Liane Lovitt, Neel Nanda, Catherine Olsson, Dario Amodei, Tom Brown, Jack Clark, Sam McCandlish, Chris Olah, Ben Mann, and Jared Kaplan.
\newblock Training a helpful and harmless assistant with reinforcement learning from human feedback.
\newblock \emph{arXiv preprint arXiv:2204.05862}, 2022{\natexlab{a}}.

\bibitem[Bai et~al.(2022{\natexlab{b}})Bai, Kadavath, Kundu, Askell, Kernion, Jones, Chen, Goldie, Azalia~Mirhoseini, Chen, Olsson, Olah, Hernandez, Drain, Ganguli, Li, Tran-Johnson, Perez, Kerr, Mueller, Ladish, Landau, Ndousse, Lukosuite, Lovitt, Sellitto, Elhage, Schiefer, Mercado, DasSarma, Lasenby, Larson, Ringer, Johnston, Kravec, Showk, Fort, Lanham, Telleen-Lawton, Tom~Conerly, Hume, Bowman, Hatfield-Dodds, Mann, Amodei, Joseph, McCandlish, Brown, and Kaplan]{bai.constitutionalai}
Yuntao Bai, Saurav Kadavath, Sandipan Kundu, Amanda Askell, Jackson Kernion, Andy Jones, Anna Chen, Anna Goldie, Cameron~McKinnon Azalia~Mirhoseini, Carol Chen, Catherine Olsson, Christopher Olah, Danny Hernandez, Dawn Drain, Deep Ganguli, Dustin Li, Eli Tran-Johnson, Ethan Perez, Jamie Kerr, Jared Mueller, Jeffrey Ladish, Joshua Landau, Kamal Ndousse, Kamile Lukosuite, Liane Lovitt, Michael Sellitto, Nelson Elhage, Nicholas Schiefer, Noemi Mercado, Nova DasSarma, Robert Lasenby, Robin Larson, Sam Ringer, Scott Johnston, Shauna Kravec, Sheer~El Showk, Stanislav Fort, Tamera Lanham, Timothy Telleen-Lawton, Tom~Henighan Tom~Conerly, Tristan Hume, Samuel~R. Bowman, Zac Hatfield-Dodds, Ben Mann, Dario Amodei, Nicholas Joseph, Sam McCandlish, Tom Brown, and Jared Kaplan.
\newblock Constitutional {AI}: {H}armlessness from {AI} feedback.
\newblock \emph{arXiv preprint arXiv:2212.08073}, 2022{\natexlab{b}}.

\bibitem[Bradley and Terry(1952)]{bradley1952rank}
Ralph~Allan Bradley and Milton~E Terry.
\newblock Rank analysis of incomplete block designs: I. the method of paired comparisons.
\newblock \emph{Biometrika}, 39\penalty0 (3/4):\penalty0 324--345, 1952.

\bibitem[Brown et~al.(2020)Brown, Mann, Ryder, Subbiah, Kaplan, Dhariwal, Neelakantan, Shyam, Sastry, Askell, et~al.]{brown2020language}
Tom~B Brown, Benjamin Mann, Nick Ryder, Melanie Subbiah, Jared Kaplan, Prafulla Dhariwal, Arvind Neelakantan, Pranav Shyam, Girish Sastry, Amanda Askell, et~al.
\newblock Language models are few-shot learners.
\newblock In \emph{Advances in neural information processing systems}, volume~33, pages 1877--1901, 2020.

\bibitem[Cheng et~al.(2023)Cheng, Xie, Bai, Dai, and Du]{dsp}
Pengyu Cheng, Jiawen Xie, Ke~Bai, Yong Dai, and Nan Du.
\newblock Everyone deserves {A} reward: Learning customized human preferences.
\newblock \emph{arXiv preprint arXiv:2309.03126}, 2023.

\bibitem[Christiano et~al.(2017)Christiano, Leike, Brown, Martic, Legg, and Amodei]{christiano2017deep}
Paul~F Christiano, Jan Leike, Tom~B Brown, Miljan Martic, Shane Legg, and Dario Amodei.
\newblock Deep reinforcement learning from human preferences.
\newblock In \emph{Advances in Neural Information Processing Systems}, pages 4299--4307, 2017.

\bibitem[Dong et~al.(2025)Dong, Dong, Tang, Ye, Sun, Sui, and Wei]{dong2025reinforcement}
Qingxiu Dong, Li~Dong, Yao Tang, Tianzhu Ye, Yutao Sun, Zhifang Sui, and Furu Wei.
\newblock Reinforcement pre-training.
\newblock \emph{arXiv preprint arXiv:2506.08007}, 2025.

\bibitem[Ethayarajh et~al.(2024)Ethayarajh, Xu, Muennighoff, Jurafsky, and Kiela]{ethayarajh2024kto}
Kawin Ethayarajh, Winnie Xu, Niklas Muennighoff, Dan Jurafsky, and Douwe Kiela.
\newblock Kto: Model alignment as prospect theoretic optimization.
\newblock \emph{arXiv preprint arXiv:2402.01306}, 2024.

\bibitem[Garg et~al.(2021)Garg, Chakraborty, Cundy, Song, and Ermon]{garg2021iq-learn}
Divyansh Garg, Shuvam Chakraborty, Chris Cundy, Jiaming Song, and Stefano Ermon.
\newblock Iq-learn: Inverse soft-q learning for imitation.
\newblock In \emph{Advances in Neural Information Processing Systems 34}, pages 4028--4039, 2021.

\bibitem[Geist et~al.(2019)Geist, Scherrer, and Pietquin]{geist2019theory}
Matthieu Geist, Bruno Scherrer, and Olivier Pietquin.
\newblock A theory of regularized markov decision processes.
\newblock In \emph{The 36th International Conference on Machine Learning}, pages 2160--2169, 2019.

\bibitem[Gu et~al.(2025)Gu, Jiang, Shi, Tan, Zhai, Xu, Li, Shen, Ma, Liu, Wang, Zhang, Wang, Gao, Ni, and Guo]{judge}
Jiawei Gu, Xuhui Jiang, Zhichao Shi, Hexiang Tan, Xuehao Zhai, Chengjin Xu, Wei Li, Yinghan Shen, Shengjie Ma, Honghao Liu, Saizhuo Wang, Kun Zhang, Yuanzhuo Wang, Wen Gao, Lionel Ni, and Jian Guo.
\newblock A survey on llm-as-a-judge.
\newblock \emph{arxiv prepring, arxiv:2411.15594}, 2025.

\bibitem[Haarnoja et~al.(2017)Haarnoja, Tang, Abbeel, and Levine]{haarnoja2017reinforcement}
Tuomas Haarnoja, Haoran Tang, Pieter Abbeel, and Sergey Levine.
\newblock Reinforcement learning with deep energy-based policies.
\newblock In \emph{The 34th International Conference on Machine Learning}, pages 1352--1361, 2017.

\bibitem[He et~al.(2024)He, Luo, Bai, Hu, Thai, Shen, Hu, Han, Huang, Zhang, et~al.]{he2024olympiadbench}
Chaoqun He, Renjie Luo, Yuzhuo Bai, Shengding Hu, Zhen~Leng Thai, Junhao Shen, Jinyi Hu, Xu~Han, Yujie Huang, Yuxiang Zhang, et~al.
\newblock Olympiadbench: A challenging benchmark for promoting agi with olympiad-level bilingual multimodal scientific problems.
\newblock \emph{CoRR}, abs/2402.14008, 2024.

\bibitem[Hendrycks et~al.(2021)Hendrycks, Burns, Kadavath, Arora, Basart, Tang, Song, and Steinhardt]{hendrycks2measuring}
Dan Hendrycks, Collin Burns, Saurav Kadavath, Akul Arora, Steven Basart, Eric Tang, Dawn Song, and Jacob Steinhardt.
\newblock Measuring mathematical problem solving with the math dataset.
\newblock In \emph{The 35th Conference on Neural Information Processing Systems Datasets and Benchmarks Track}, 2021.

\bibitem[Hinton(2015)]{hinton2015distilling}
Geoffrey Hinton.
\newblock Distilling the knowledge in a neural network.
\newblock \emph{CoRR}, abs/1503.02531, 2015.

\bibitem[Lee et~al.(2024{\natexlab{a}})Lee, Phatale, Mansoor, Mesnard, Ferret, Lu, Bishop, Hall, Carbune, Rastogi, and Prakash]{RLAIF}
Harrison Lee, Samrat Phatale, Hassan Mansoor, Thomas Mesnard, Johan Ferret, Kellie Lu, Colton Bishop, Ethan Hall, Victor Carbune, Abhinav Rastogi, and Sushant Prakash.
\newblock {RLAIF} vs. {RLHF}: {S}caling reinforcement learning from human feedback with {AI} feedback.
\newblock In \emph{Proceedings of the 41st International Conference on Machine Learning (ICML)}, 2024{\natexlab{a}}.

\bibitem[Lee et~al.(2024{\natexlab{b}})Lee, Park, Kim, and Seo]{mf-bench}
Seongyun Lee, Sue~Hyun Park, Seungone Kim, and Minjoon Seo.
\newblock Aligning to thousands of preferences via system message generalization.
\newblock In \emph{Advances in Neural Information Processing Systems (NeurIPS)}, 2024{\natexlab{b}}.

\bibitem[Lewkowycz et~al.(2022)Lewkowycz, Andreassen, Dohan, Dyer, Michalewski, Ramasesh, Slone, Anil, Schlag, Gutman{-}Solo, Wu, Neyshabur, Gur{-}Ari, and Misra]{DBLP:journals/corr/abs-2206-14858}
Aitor Lewkowycz, Anders Andreassen, David Dohan, Ethan Dyer, Henryk Michalewski, Vinay~V. Ramasesh, Ambrose Slone, Cem Anil, Imanol Schlag, Theo Gutman{-}Solo, Yuhuai Wu, Behnam Neyshabur, Guy Gur{-}Ari, and Vedant Misra.
\newblock Solving quantitative reasoning problems with language models.
\newblock \emph{CoRR}, abs/2206.14858, 2022.

\bibitem[Li et~al.(2024{\natexlab{a}})Li, Zeng, Liao, Li, Kang, Garcia, and Hong]{li2024joint}
Chenliang Li, Siliang Zeng, Zeyi Liao, Jiaxiang Li, Dongyeop Kang, Alfredo Garcia, and Mingyi Hong.
\newblock Joint demonstration and preference learning improves policy alignment with human feedback.
\newblock \emph{arXiv preprint arXiv:2406.06874}, 2024{\natexlab{a}}.

\bibitem[Li et~al.(2024{\natexlab{b}})Li, Beeching, Tunstall, Lipkin, Soletskyi, Huang, Rasul, Yu, Jiang, Shen, et~al.]{li2024numinamath}
Jia Li, Edward Beeching, Lewis Tunstall, Ben Lipkin, Roman Soletskyi, Shengyi Huang, Kashif Rasul, Longhui Yu, Albert~Q Jiang, Ziju Shen, et~al.
\newblock Numinamath: The largest public dataset in ai4maths with 860k pairs of competition math problems and solutions.
\newblock \emph{Hugging Face repository}, page~9, 2024{\natexlab{b}}.

\bibitem[Liu et~al.(2025)Liu, Yao, Min, Cao, Hou, and Li]{rm-bench}
Yantao Liu, Zijun Yao, Rui Min, Yixin Cao, Lei Hou, and Juanzi Li.
\newblock {RM-Bench}: Benchmarking reward models of language models with subtlety and style.
\newblock In \emph{International Conference on Learning Representations (ICLR)}, 2025.

\bibitem[Mahan et~al.(2024)Mahan, Van~Phung, Rafailov, Blagden, Lile, Castricato, Fr{\"a}nken, Finn, and Albalak]{mahan2024generative}
Dakota Mahan, Duy Van~Phung, Rafael Rafailov, Chase Blagden, Nathan Lile, Louis Castricato, Jan-Philipp Fr{\"a}nken, Chelsea Finn, and Alon Albalak.
\newblock Generative reward models.
\newblock \emph{arXiv preprint arXiv:2410.12832}, 2024.

\bibitem[Malik et~al.(2025)Malik, Pyatkin, Land, Morrison, Smith, Hajishirzi, and Lambert]{malik2025rewardbench}
Saumya Malik, Valentina Pyatkin, Sander Land, Jacob Morrison, Noah~A Smith, Hannaneh Hajishirzi, and Nathan Lambert.
\newblock Rewardbench 2: Advancing reward model evaluation.
\newblock \emph{arXiv preprint arXiv:2506.01937}, 2025.

\bibitem[Ng and Russell(2000)]{ng2000algorithms}
Andrew~Y. Ng and Stuart~J. Russell.
\newblock Algorithms for inverse reinforcement learning.
\newblock In \emph{Proceedings of the Seventeenth International Conference on Machine Learning (ICML 2000)}, pages 663--670, 2000.

\bibitem[Ng et~al.(1999)Ng, Harada, and Russell]{ng1999policy}
Andrew~Y Ng, Daishi Harada, and Stuart Russell.
\newblock Policy invariance under reward transformations: Theory and application to reward shaping.
\newblock In \emph{The 16th International Conference on Machine Learning}, pages 278--287, 1999.

\bibitem[OpenAI(2022)]{openai2022chatgpt}
OpenAI.
\newblock Introducing chatgpt.
\newblock \emph{https://openai.com/blog/chatgpt}, 2022.

\bibitem[Osa et~al.(2018)Osa, Pajarinen, Neumann, Bagnell, Abbeel, and Peters]{osa2018survey}
Takayuki Osa, Joni Pajarinen, Gerhard Neumann, J.~Andrew Bagnell, Pieter Abbeel, and Jan Peters.
\newblock An algorithmic perspective on imitation learning.
\newblock \emph{Foundations and Trends in Robotic}, 7\penalty0 (1-2):\penalty0 1--179, 2018.

\bibitem[Ouyang et~al.(2022)Ouyang, Wu, Jiang, Almeida, Wainwright, Mishkin, Zhang, Agarwal, Slama, Ray, Schulman, Hilton, Kelton, Miller, Simens, Askell, Welinder, Christiano, Leike, and Lowe]{trainlmfollowinstruct}
Long Ouyang, Jeffrey Wu, Xu~Jiang, Diogo Almeida, Carroll~L. Wainwright, Pamela Mishkin, Chong Zhang, Sandhini Agarwal, Katarina Slama, Alex Ray, John Schulman, Jacob Hilton, Fraser Kelton, Luke Miller, Maddie Simens, Amanda Askell, Peter Welinder, Paul~F. Christiano, Jan Leike, and Ryan Lowe.
\newblock Training language models to follow instructions with human feedback.
\newblock In \emph{Advances in Neural Information Processing Systems 36}, 2022.

\bibitem[Pang et~al.(2023)Pang, Wang, Li, Chen, Xu, and Zhang]{pang2023yang}
Jing-Cheng Pang, Pengyuan Wang, Kaiyuan Li, Xiong-Hui Chen, Jiacheng Xu, and Yang~Yu. Zhang, Zongzhang.
\newblock Language model self-improvement by reinforcement learning contemplation.
\newblock In \emph{International Conference on Learning Representations}, 2023.

\bibitem[Pomerleau(1991)]{Pomerleau91bc}
Dean Pomerleau.
\newblock Efficient training of artificial neural networks for autonomous navigation.
\newblock \emph{Neural Computation}, 3\penalty0 (1):\penalty0 88--97, 1991.

\bibitem[Radford et~al.(2019)Radford, Wu, Child, Luan, Amodei, Sutskever, et~al.]{radford2019language}
Alec Radford, Jeffrey Wu, Rewon Child, David Luan, Dario Amodei, Ilya Sutskever, et~al.
\newblock Language models are unsupervised multitask learners.
\newblock \emph{OpenAI blog}, 1\penalty0 (8):\penalty0 9, 2019.

\bibitem[Rafailov et~al.(2023)Rafailov, Sharma, Mitchell, Ermon, Manning, and Finn]{rafailov2023direct}
Rafael Rafailov, Archit Sharma, Eric Mitchell, Stefano Ermon, Christopher~D Manning, and Chelsea Finn.
\newblock Direct preference optimization: Your language model is secretly a reward model.
\newblock \emph{arXiv preprint arXiv:2305.18290}, 2023.

\bibitem[Rajaraman et~al.(2020)Rajaraman, Yang, Jiao, and Ramchandran]{rajaraman2020fundamental}
Nived Rajaraman, Lin~F. Yang, Jiantao Jiao, and Kannan Ramchandran.
\newblock Toward the fundamental limits of imitation learning.
\newblock In \emph{Advances in Neural Information Processing Systems 33}, pages 2914--2924, 2020.

\bibitem[Ramamurthy et~al.(2023)Ramamurthy, Ammanabrolu, Brantley, Hessel, Sifa, Bauckhage, Hajishirzi, and Choi]{llm_formalization}
Rajkumar Ramamurthy, Prithviraj Ammanabrolu, Kiant{\'{e}} Brantley, Jack Hessel, Rafet Sifa, Christian Bauckhage, Hannaneh Hajishirzi, and Yejin Choi.
\newblock Is reinforcement learning (not) for natural language processing: Benchmarks, baselines, and building blocks for natural language policy optimization.
\newblock In \emph{International Conference on Learning Representations (ICLR)}, 2023.

\bibitem[Ross and Bagnell(2010)]{ross2010efficient}
St{\'e}phane Ross and Drew Bagnell.
\newblock Efficient reductions for imitation learning.
\newblock In \emph{Proceedings of the 13rd International Conference on Artificial Intelligence and Statistics}, pages 661--668, 2010.

\bibitem[Shao et~al.(2024)Shao, Wang, Zhu, Xu, Song, Bi, Zhang, Zhang, Li, Wu, et~al.]{shao2024deepseekmath}
Zhihong Shao, Peiyi Wang, Qihao Zhu, Runxin Xu, Junxiao Song, Xiao Bi, Haowei Zhang, Mingchuan Zhang, YK~Li, Y~Wu, et~al.
\newblock Deepseekmath: Pushing the limits of mathematical reasoning in open language models.
\newblock \emph{CoRR}, abs/2402.03300, 2024.

\bibitem[Sun et~al.(2023)Sun, H{\"u}y{\"u}k, and van~der Schaar]{sun2023query}
Hao Sun, Alihan H{\"u}y{\"u}k, and Mihaela van~der Schaar.
\newblock Query-dependent prompt evaluation and optimization with offline inverse rl.
\newblock \emph{arXiv preprint arXiv:2309.06553}, 2023.

\bibitem[Vaswani(2017)]{vaswani2017attention}
A~Vaswani.
\newblock Attention is all you need.
\newblock \emph{Advances in Neural Information Processing Systems 31}, 2017.

\bibitem[Wulfmeier et~al.(2024)Wulfmeier, Bloesch, Vieillard, Ahuja, Bornschein, Huang, Sokolov, Barnes, Desjardins, Bewley, et~al.]{wulfmeier2024imitating}
Markus Wulfmeier, Michael Bloesch, Nino Vieillard, Arun Ahuja, Jorg Bornschein, Sandy Huang, Artem Sokolov, Matt Barnes, Guillaume Desjardins, Alex Bewley, et~al.
\newblock Imitating language via scalable inverse reinforcement learning.
\newblock \emph{Advances in Neural Information Processing Systems 37}, pages 90714--90735, 2024.

\bibitem[Xu et~al.(2020)Xu, Li, and Yu]{xu2020error}
Tian Xu, Ziniu Li, and Yang Yu.
\newblock Error bounds of imitating policies and environments.
\newblock In \emph{Advances in Neural Information Processing Systems 33}, pages 15737--15749, 2020.

\bibitem[Xu et~al.(2021)Xu, Li, and Yu]{xu2021error}
Tian Xu, Ziniu Li, and Yang Yu.
\newblock Error bounds of imitating policies and environments for reinforcement learning.
\newblock \emph{IEEE Transactions on Pattern Analysis and Machine Intelligence}, 2021.

\bibitem[Xu et~al.(2022)Xu, Li, Yu, and Luo]{xu2022understanding}
Tian Xu, Ziniu Li, Yang Yu, and Zhi-Quan Luo.
\newblock Understanding adversarial imitation learning in small sample regime: A stage-coupled analysis.
\newblock \emph{arXiv}, 2208.01899, 2022.

\bibitem[Yang et~al.(2024)Yang, Yang, Zhang, Hui, Zheng, Yu, Li, Liu, Huang, Wei, Lin, Yang, Tu, Zhang, Yang, Yang, Zhou, Lin, Dang, Lu, Bao, Yang, Yu, Li, Xue, Zhang, Zhu, Men, Lin, Li, Xia, Ren, Ren, Fan, Su, Zhang, Wan, Liu, Cui, Zhang, and Qiu]{qwen2.5}
An~Yang, Baosong Yang, Beichen Zhang, Binyuan Hui, Bo~Zheng, Bowen Yu, Chengyuan Li, Dayiheng Liu, Fei Huang, Haoran Wei, Huan Lin, Jian Yang, Jianhong Tu, Jianwei Zhang, Jianxin Yang, Jiaxi Yang, Jingren Zhou, Junyang Lin, Kai Dang, Keming Lu, Keqin Bao, Kexin Yang, Le~Yu, Mei Li, Mingfeng Xue, Pei Zhang, Qin Zhu, Rui Men, Runji Lin, Tianhao Li, Tingyu Xia, Xingzhang Ren, Xuancheng Ren, Yang Fan, Yang Su, Yichang Zhang, Yu~Wan, Yuqiong Liu, Zeyu Cui, Zhenru Zhang, and Zihan Qiu.
\newblock {Qwen2.5} technical report.
\newblock \emph{arXiv preprint arXiv:2412.15115}, 2024.

\bibitem[Yu et~al.(2025)Yu, Zeng, Gu, Wang, Wang, Meng, Zhou, Zhang, Zhang, and Ye]{yu2025rewardanything}
Zhuohao Yu, Jiali Zeng, Weizheng Gu, Yidong Wang, Jindong Wang, Fandong Meng, Jie Zhou, Yue Zhang, Shikun Zhang, and Wei Ye.
\newblock Rewardanything: Generalizable principle-following reward models.
\newblock \emph{arXiv preprint arXiv:2506.03637}, 2025.

\bibitem[Zhang et~al.(2024)Zhang, Hosseini, Bansal, Kazemi, Kumar, and Agarwal]{zhang2024generative}
Lunjun Zhang, Arian Hosseini, Hritik Bansal, Mehran Kazemi, Aviral Kumar, and Rishabh Agarwal.
\newblock {Generative Verifiers}: Reward modeling as next-token prediction.
\newblock \emph{CoRR}, abs/2408.15240, 2024.

\bibitem[Zhao et~al.(2025)Zhao, Kang, Feng, Levine, and Song]{zhao2025learning}
Xuandong Zhao, Zhewei Kang, Aosong Feng, Sergey Levine, and Dawn Song.
\newblock Learning to reason without external rewards.
\newblock \emph{arXiv preprint arXiv:2505.19590}, 2025.

\bibitem[Zheng et~al.(2023)Zheng, Chiang, Sheng, Zhuang, Wu, Zhuang, Lin, Li, Li, Xing, et~al.]{zheng2023judging}
Lianmin Zheng, Wei-Lin Chiang, Ying Sheng, Siyuan Zhuang, Zhanghao Wu, Yonghao Zhuang, Zi~Lin, Zhuohan Li, Dacheng Li, Eric Xing, et~al.
\newblock Judging llm-as-a-judge with mt-bench and chatbot arena.
\newblock \emph{arXiv preprint arXiv:2306.05685}, 2023.

\bibitem[Zhou and Jiang(2004)]{zhou2004nec45}
Zhi-Hua Zhou and Yuan Jiang.
\newblock {NeC4.5}: Neural ensemble based {C4.5}.
\newblock \emph{IEEE Transactions on Knowledge and Data Engineering}, 16\penalty0 (6):\penalty0 770--773, 2004.

\bibitem[Ziebart et~al.(2008)Ziebart, Maas, Bagnell, and Dey]{ziebart2008maximum}
Brian~D. Ziebart, Andrew~L. Maas, J.~Andrew Bagnell, and Anind~K. Dey.
\newblock Maximum entropy inverse reinforcement learning.
\newblock In \emph{Proceedings of the AAAI conference on artificial intelligence}, pages 1433--1438, 2008.

\end{thebibliography}
